\documentclass{article}
\usepackage[utf8]{inputenc}
\usepackage{authblk}
\usepackage{setspace}
\usepackage[margin=1.25in]{geometry}
\usepackage{graphicx}
\usepackage{subcaption}
\usepackage{amsmath}
\usepackage{amssymb}
\usepackage{mathtools}
\usepackage{amsfonts}
\usepackage{amsthm}
\usepackage{soul}
\usepackage[
    style=nejm,
    citestyle=numeric-comp,
    sorting=none
]{biblatex}
\usepackage{amsmath,amssymb}
\usepackage[ruled,vlined,linesnumbered]{algorithm2e}
\SetKwInput{KwInput}{Input}
\SetKwInput{KwOutput}{Output}
\DontPrintSemicolon
\usepackage{rotating}

\usepackage{tabularx}

\usepackage{array}

\usepackage{booktabs}
\usepackage{multirow}
\usepackage{threeparttable}
\usepackage{pdflscape}  
\usepackage{adjustbox}  
\usepackage{xcolor,colortbl}
\usepackage{makecell}
\usepackage{color}
\usepackage{tikz}

\usepackage[pagebackref,breaklinks=true,colorlinks,citecolor=blue,urlcolor=magenta,linkcolor=blue,bookmarks=false,backref=false]{hyperref}

\graphicspath{{./}}
\title{A Granularity-Aware EEG Feature Framework for Psychopathology Dimension Prediction}

\author[1$\dag$]{Haofan Cheng}
\author[1$\dag$]{Jingjing Hu}
\author[1]{Jingrong Pei}
\author[1]{Shuaiqi Fu}
\author[3]{Meilun Shen}
\author[1]{Shuai Fang}
\author[1,2]{Meng Wang}
\author[1,2*]{Dan Guo}
\author[3*]{Jie Zhang}

\affil[1]{School of Computer Science and Information Engineering, Hefei University of Technology, Hefei, 230601, China}
\affil[2]{Key Laboratory of Knowledge Engineering with Big Data (Ministry of Education of China), Hefei University of Technology, Hefei, 230601, China}
\affil[3]{Preventive Medicine Institute \& Medical Innovation Center, The Fourth Military Medical University, Xi'an, 710032, China}
\affil[*]{Address correspondence to: guodan@hfut.edu.cn; zhangjie78@fmmu.edu.cn}
\affil[$\dag$]{These authors contributed equally to this work.}

\date{}

\begin{document}

\maketitle

\begin{abstract}
Electroencephalography (EEG) offers a noninvasive approach for examining neurophysiological correlates of dimensional psychopathology, yet systematic evidence across EEG paradigms and feature granularities remains limited. Here, we develop a granularity-aware EEG feature pipeline that organizes multi-scale descriptors into global, regional, and channel levels. Using the Healthy Brain Network (HBN) cohort, we evaluate prediction of four psychopathology dimensions—$p$-factor, internalizing, externalizing, and attention problems—across four EEG paradigms. Given the heterogeneity of pediatric psychopathology and the moderate reliability of questionnaire-derived scores, this setting represents a challenging feasibility test rather than a clinical screening scenario. Tree-based models and granularity-balanced feature selection showed promising improvements over conventional approaches in certain conditions, though effect sizes remained modest. Visualization of selected markers identified dimension-specific spatial and spectral patterns aligned with existing neurophysiological knowledge. An exploratory cross-dataset sanity check on the independent PEARL cohort suggested that the selection principle remains technically operable under protocol shifts, without claiming generalizability. Overall, multi-scale EEG features contain weak but detectable signals related to dimensional psychopathology, and granularity-aware selection may serve as a useful feature-reduction strategy for future EEG-based phenotyping studies.
\end{abstract}


\section*{Introduction}
Mental health problems in children and adolescents have become an increasingly important public health concern, yet emotional, behavioral, and attentional vulnerabilities often emerge in heterogeneous and context-dependent forms, making early identification particularly challenging. Current practice still largely depends on self-report questionnaires, parent- or teacher-rated scales, and clinical interviews. Although indispensable, these tools are susceptible to recall bias, social desirability, rater inconsistency, and limited sensitivity to subclinical or early-stage changes, driving precision psychiatry toward objective and individualized neurobiological markers beyond symptom-level observation~\cite{bosl2025dynamical,14comai2025moving}. A growing body of evidence suggests that psychopathology is better characterized as continuous dimensions—such as the general $p$-factor, internalizing, externalizing, and attention problems—rather than strictly separated diagnostic categories. While these dimensional frameworks offer a useful bridge between clinical symptoms and underlying neural mechanisms, they remain primarily derived from behavioral rating scales, leaving a critical gap in objective neurophysiological validation, particularly in pediatric populations where functional brain network organization relates to cognitive variability~\cite{new_11}.

Electroencephalography (EEG) provides a promising tool for addressing this gap. EEG is non-invasive, relatively low-cost, and capable of capturing neural activity with millisecond-level temporal resolution, making it especially suitable for large-scale pediatric studies. Recent reviews have highlighted the potential of EEG-based biomarkers for psychiatric disorders, while also noting challenges in robustness, heterogeneity, and clinical translation~\cite{2yun2024advances}. Recent advances in brain--computer interface and neurophysiological signal analysis have further demonstrated the potential of EEG for objective cognitive and affective assessment~\cite{jia2024enhancing,ju2022recognition}. For example, adaptive spatiotemporal encoding networks have been developed for cognitive assessment from resting-state EEG~\cite{new_2}, and resting EEG has also been explored for decoding individual psychological traits~\cite{4jach2020decoding}. EEG-based modeling has shown effectiveness in emotion recognition~\cite{ye2024adaptive,li2024domain,si2023cross}, driving state detection~\cite{qi2024augmented}, and moral cognition analysis~\cite{bao2023predicting}. Beyond EEG, related studies on subtle behavioral and affective cues, such as micro-action recognition, further highlight the importance of objective multimodal markers for understanding human psychological and behavioral states~\cite{guo2024benchmarking}. More recently, EEG emotion recognition has benefited from advanced representation learning strategies, including facial emoji proxy modeling~\cite{hu2026see} and spatial-energy-aware dynamic filtering with sparse graph convolutions~\cite{hu2025spatialenergyaware,taoband}, suggesting that structured EEG representations can capture meaningful affective and cognitive information. 

Despite these advances, several challenges remain in EEG-based psychopathology prediction. First, most existing studies focus on a single symptom domain, clinical label, or psychological construct, while systematic prediction across multiple psychopathology dimensions remains limited. Second, many studies rely on task-specific EEG paradigms or relatively small samples, which may reduce generalizability across recording conditions and developmental populations. Resting-state EEG is attractive because it is easy to administer, task-free, and less affected by performance-related confounds; however, task-state paradigms may also contain complementary information related to naturalistic audiovisual processing or visual contextual modulation. Therefore, more comparison across multiple EEG paradigms is needed. Third, although previous work has explored data augmentation and signal processing strategies to improve model robustness under limited-sample settings~\cite{dong2023approach,wang2025enabling}, the organization of EEG features across spatial scales remains underexplored. Recent work on multidimensional EEG feature integration and feature selection has shown that combining heterogeneous EEG descriptors can improve precision diagnosis in psychiatric contexts~\cite{5luo2025multidimensional}. However, conventional feature selection strategies often pool all features together, which may allow high-dimensional channel-level descriptors to dominate the selected feature set and suppress lower-dimensional but potentially stable global or regional markers. This is particularly problematic because different spatial granularities capture distinct neural phenomena and exhibit inherent trade-offs: channel-level features preserve fine-grained topographical details but are sensitive to anatomical variability, while global features offer robustness at the cost of potentially underfitting symptom-specific patterns. Existing methods rarely account for such multi-scale conflicts, leaving a gap in principled strategies for balanced feature representation.

Another important issue is the reliability and interpretability of EEG-based computational pipelines. Neuroimaging and neurophysiological studies have increasingly emphasized transparent and reproducible analytical practices~\cite{3poldrack2017scanning}. In medical AI, performance-oriented models may also suffer from limited explainability, which can hinder clinical trust and practical deployment~\cite{6london2019artificial}. These concerns are particularly important in pediatric psychopathology prediction, where models should not only achieve predictive performance but also provide stable and interpretable evidence about which frequency bands, brain regions, and spatial granularities contribute to different psychopathology dimensions.

To address these limitations, we propose a granularity-aware EEG feature framework for predicting dimensional psychopathology scores from multi-paradigm EEG recordings. The core idea is to treat global, regional, and channel-level descriptors not as interchangeable candidates but as separate information pools, each capturing complementary yet potentially conflicting aspects of neural activity. Rather than allowing high-dimensional channel features to dominate via flat selection, our framework enforces a balanced multi-scale representation by allocating independent feature budgets to each granularity. Using the large-scale Healthy Brain Network (HBN) cohort as the primary benchmark, we investigate four EEG paradigms: resting-state eyes closed, resting-state eyes open, movie watching, and surround suppression. For each paradigm, we extract a multi-scale feature set covering spectral power, band-power ratios, differential entropy, aperiodic activity, signal complexity, microstate dynamics, regional asymmetry, connectivity, and temporal dynamics. These features are organized into three spatial granularities: global, region, and channel. We then systematically evaluate their predictive capacity for four psychopathology dimensions, including $p$-factor, internalizing, externalizing, and attention problems. To further examine the generalizability of the proposed framework beyond the HBN pediatric cohort, we additionally conduct an exploratory cross-dataset assessment on the PEARL dataset~\cite{dzianok2024pearl}, where resting-state EEG is used to predict clinically relevant depression and neuroticism-related measures under the same feature extraction and granularity-aware selection pipeline. As part of this evaluation, we compare GSTS against multiple feature selection strategies across all four EEG paradigms (Figure~\ref{fig:ec_mw_comparison}), providing empirical evidence on the relative strengths and limitations of granularity-aware selection in different recording conditions.

The main contributions of this study are summarized as follows:
\begin{enumerate}
    \item We establish a large-scale multi-paradigm evaluation protocol using the HBN dataset, systematically comparing four EEG paradigms and four psychopathology dimensions to assess the feasibility of EEG-based prediction in pediatric populations.

    \item We propose a granularity-aware stability selection strategy (GSTS) that performs feature selection separately across global, regional, and channel-level descriptors. Comparative experiments (Figure~\ref{fig:ec_mw_comparison}) and visualization of selected markers (Figure~\ref{fig:interpretable_markers}) demonstrate its competitive performance and interpretability.

    \item We implement a fold-contained evaluation pipeline integrating model screening, feature selection, hyperparameter tuning, and ablation analysis, with all operations confined to training folds for unbiased generalization estimates, and conduct an exploratory cross-dataset assessment on the PEARL cohort to examine transferability under protocol shifts.
\end{enumerate}

\begin{figure}[t]
    \centering
    \includegraphics[width=1\linewidth]{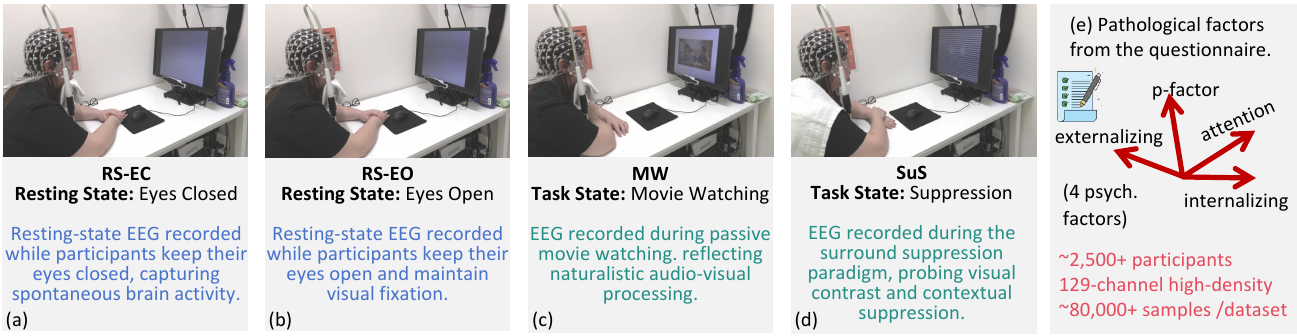}
    \caption{\textbf{Overview of the experimental datasets.} (a) Resting-state with eyes closed (RS-EC). (b) Resting-state with eyes open (RS-EO). (c) Passive movie watching (MW). (d) Surround suppression (SuS). (e) Four psychopathology dimension scores derived from the Child Behavior Checklist (CBCL) via a bifactor model: $p$-factor, internalizing, externalizing, and attention. The dataset comprises approximately 2,600 participants aged 5--21 years, with 129-channel high-density EEG recordings and {over 80,000 4-s epochs per paradigm.}}
    \label{fig:task}
\end{figure}

\section*{Materials and Methods} \label{sec:Materials_and_Methods}

\begin{figure}[t]
    \centering
    \includegraphics[width=1\linewidth]{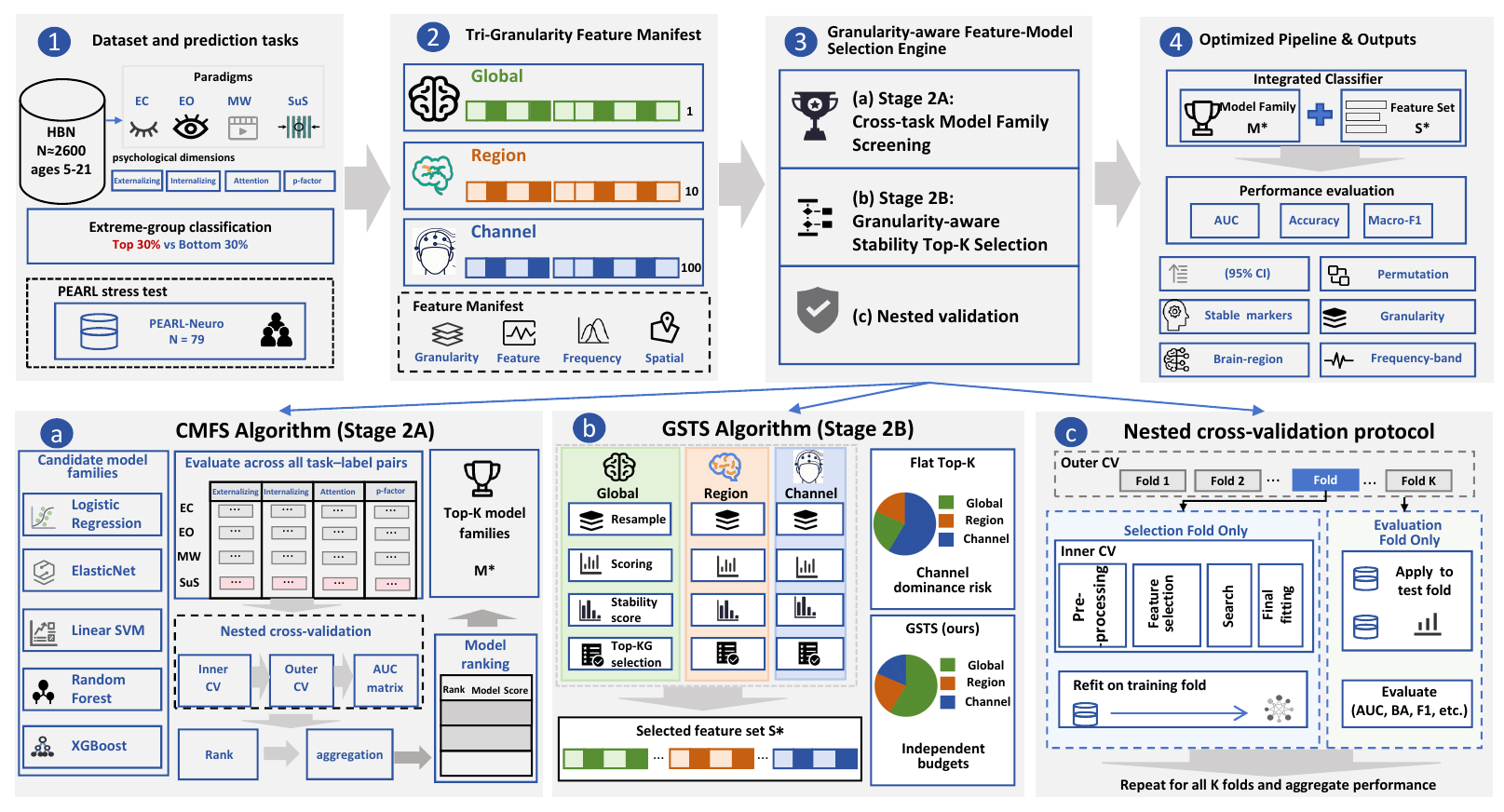}
    \caption{\textbf{Overview of the proposed framework.} The pipeline consists of four stages: (1) dataset preparation from the HBN cohort with four EEG tasks and four psychopathology dimensions; (2) a granularity-aware feature-model selection engine comprising cross-task model family screening (CMFS), granularity-aware stability-based feature selection (GSTS), and {fold-contained} nested cross-validation; (3) an optimized multi-granularity pipeline with performance evaluation and interpretable outputs; and (4) a {fold-contained} nested cross-validation protocol ensuring unbiased generalization estimates across candidate model families (Logistic Regression, ElasticNet, Linear SVM, Random Forest, XGBoost).}
    \label{fig:main}
\end{figure}

\begin{table}[t]
    \centering
    \small
    \caption{\textbf{Summary of four EEG paradigms from the HBN dataset.} Each epoch is a 4\,s segment at 100\,Hz over 129 channels. SuS uses a block-design SSVEP paradigm ($\sim$5--10 min/subject), yielding $\sim$10$\times$ more epochs per subject than resting-state ($\sim$2.5 min) and MW ($\sim$2--3 min).}
    \label{tab:dataset_summary}
    \renewcommand{\arraystretch}{1.15}
    \resizebox{0.7\linewidth}{!}{%
    \begin{tabular}{l c c c c}
        \toprule
        \textbf{Task} & \textbf{\#Subject} & \textbf{\#Epochs} & \textbf{Epochs/Subj.} & \textbf{Description} \\
        \midrule
        EC (Eyes Closed) & 2,575 & 82,814 & 32.2 & Resting-state, eyes closed \\
        EO (Eyes Open)   & 2,592 & 85,521 & 33.0 & Resting-state, eyes open \\
        MW (Movie Watching) & 2,295 & 88,258 & 38.5 & Naturalistic audiovisual processing \\
        SuS (Surround Suppression) & 1,925 & 780,587 & 405.5 & Visual contrast integration, run-1 \\
        \bottomrule
    \end{tabular}
    }
\end{table}

\subsection*{Experimental Datasets}  \label{sec:Experimental_Datasets}

We utilized four EEG paradigms from the Healthy Brain Network (HBN) dataset~\cite{shirazi2024hbn}, a large-scale community-based study of psychiatric and learning disorders in children and adolescents aged 5--21 years (approved by the Chesapeake IRB), employing the V2 processed version from Releases 1--11. These four paradigms include two resting-state tasks—eyes closed (EC) and eyes open (EO)—and two naturalistic/task-based paradigms—movie watching (MW, Despicable Me clip) and surround suppression (SuS, run-1, block-design SSVEP, $\sim$5--10 min/subject). After task-specific quality control, the available sample sizes were 2,575 participants for EC, 2,592 for EO, 2,295 for MW, and 1,925 for SuS. Because task availability and data quality differed across participants, these samples were partially overlapping rather than identical subsets. EEG was recorded using a 129-channel EGI GSN-HydroCel system at 1,000 Hz, downsampled to 100 Hz, with 4-second non-overlapping epochs extracted from each task. The primary outcomes are four psychopathology dimension scores derived from the Child Behavior Checklist (CBCL) via a bifactor model: $p$-factor, attention, internalizing, and externalizing, all z-scored across the full cohort (2,541 participants had complete labels; the remaining 34 were excluded from supervised analyses). As summarized in Table~\ref{tab:dataset_summary} and illustrated in Figure~\ref{fig:task}, the SuS task yielded $\sim$10$\times$ more epochs per subject than resting-state ($\sim$2.5 min) and MW ($\sim$2--3 min). After requiring non-missing psychometric labels, the extreme-quantile (30/70) transform yielded task-specific binary classification sample sizes of approximately 1,527 (EC), 1,537 (EO), 1,232 (MW), and 1,084 (SuS) subjects per label.

Given the passive EEG paradigms, the questionnaire-derived nature of the target dimensions, and the broad developmental age range, the prediction tasks examined here are inherently difficult. We therefore emphasize relative improvements over chance-level performance and consistency across paradigms rather than direct comparison with binary diagnostic classification benchmarks.

\begin{table}[t]
    \centering
    \small
    \caption{\textbf{Summary of EEG features stratified by spatial granularity.}
    Under the default configuration, $C=129$, $B_{\mathrm{freq}}=5$, $R=5$ (six regions for
    connectivity, including the auricular--mastoid reference), $P=\binom{6}{2}=15$,
    $M=2$, $Q=3$, and $K_{\mathrm{ms}}=4$, yielding 28 global,
    400 regional, and 3096 channel-level features, for a total of \textbf{3524
    conceptual features}. Each conceptual feature is aggregated across
    per-subject epochs via three statistics (mean, std, median), yielding
    $3524 \times 3 = 10{,}572$ columns in the final feature matrix. }
    \label{tab:feature_summary}
    \renewcommand{\arraystretch}{1.2}
    \resizebox{1\linewidth}{!}{%
    \begin{tabular}{l l r l}
        \toprule
        \textbf{Granularity} & \textbf{Symbol} & \textbf{Dim} & \textbf{Description \&
    Primary Function (References)} \\
        \midrule
        \multirow{4}{*}{Global}
            & $M_{k}^{\mathrm{cov}}, M_{k}^{\mathrm{occ}}, M_{k}^{\mathrm{dur}}$
            & $3K_{\mathrm{ms}}=12$ & Microstate coverage, occurrence, and mean duration for state $k$. \\
            & & & \textbf{Function:} Whole-scalp global microstate dynamics
    \cite{Michel2018,Koenig2002} \\
            & $T_{k\to l}$
            & $K_{\mathrm{ms}}^2=16$ & Microstate transition probability from state $k$ to state $l$. \\
            & & & \textbf{Function:} Temporal segmentation and transition patterns of global
    brain states \cite{Lehmann1987} \\
            & & \textit{Subtotal: 28} \\

        \midrule
        \multirow{8}{*}{Region}
            & $P_{\mathrm{abs},r,b}, P_{\mathrm{rel},r,b}$
            & $2RB=50$ & Region-mean log10 absolute and relative power in region $r$ and
    band $b$. \\
            & & & \textbf{Function:} Regional oscillatory energy and frequency dominance
    \cite{Welch1967,Buzsaki2006,Cohen2014} \\
            & $A^{\mathrm{pow}}_{r,b,m}$
            & $2RB=50$ & Inter-hemispheric asymmetry of region-level log10 absolute and
    relative power, $m \in \{\mathrm{logabs}, \mathrm{rel}\}$. \\
            & & & \textbf{Function:} Regional functional lateralization of spectral activity
    \cite{Davidson1998,Allen2004} \\
            & $\mathrm{Coh}_{r_1,r_2,b}, \mathrm{PLV}_{r_1,r_2,b}$
            & $2PB=150$ & Coherence and PLV between regional pair $(r_1,r_2)$ in band $b$
    (six regions, $P=15$). \\
            & & & \textbf{Function:} Inter-regional functional synchronization and phase
    consistency \cite{Lachaux1999,Nolte2004} \\
            & $\mu_{t,r,b,m}, \sigma_{t,r,b,m}, s_{t,r,b,m}$
            & $RBMQ=150$ & Sliding-window mean, std, and temporal slope of regional spectral
    power, $m \in \{\mathrm{logabs}, \mathrm{rel}\}$. \\
            & & & \textbf{Function:} Regional time-varying neural dynamics
    \cite{Vidaurre2018} \\
            & & \textit{Subtotal: 400} & \\
        \midrule
        \multirow{12}{*}{Channel}
            & $P_{\mathrm{abs},ch,b}$
            & $CB=645$ & Log10 absolute power at channel $ch$ and band $b$. \\
            & & & \textbf{Function:} Fine-grained spectral energy topography
    \cite{Welch1967,Buzsaki2006} \\
            & $P_{\mathrm{rel},ch,b}$
            & $CB=645$ & Relative power at channel $ch$ and band $b$. \\
            & & & \textbf{Function:} Fine-grained frequency dominance across canonical bands
    \cite{Cohen2014,Bastos2016} \\
            & $\mathrm{DE}_{ch,b}$
            & $CB=645$ & Differential entropy of band-pass filtered EEG at channel $ch$ and
    band $b$. \\
            & & & \textbf{Function:} Log-energy-like spectral information content
    \cite{Duan2013} \\
            & $R_{\theta/\beta,ch}, R_{\alpha/\beta,ch}, R_{\theta/\alpha,ch}$
            & $3C=387$ & Log10 band-power ratios at channel $ch$. \\
            & & & \textbf{Function:} Channel-level spectral balance across canonical bands
    \cite{Monastra2001,Clarke2001,Arns2013} \\
            & $\kappa_{ch}, \omega_{ch}$
            & $2C=258$ & Aperiodic $1/f$ slope and intercept at channel $ch$. \\
            & & & \textbf{Function:} Channel-level scale-free neural activity and background
    characteristics \cite{He2010,Donoghue2020} \\
            & $H_{a,ch}, H_{m,ch}, H_{c,ch}, H_{\mathrm{SE},ch}$
            & $4C=516$ & Hjorth activity, mobility, complexity, and spectral entropy at
    channel $ch$. \\
            & & & \textbf{Function:} Channel-level signal complexity and irregularity
    \cite{Hjorth1970,liang2015eeg} \\
            & & \textit{Subtotal: 3096} & \\
        \midrule
            & \textbf{Total} & \textbf{3524} & Conceptual features (28 global + 400 regional +
    3096 channel) \\
        \bottomrule
    \end{tabular}%
    }
\end{table}

\subsection*{Feature Extraction} \label{sec:Feature_Extraction}

Guided by the multi-scale conflict discussed above, we extracted EEG features organized into three spatial granularities—global, region, and channel—to preserve their distinct neural information as separate candidate pools rather than treating all descriptors interchangeably. Before detailing the specific feature families, we formally define what we mean by a \textit{granularity-aware} feature selection pipeline. Let $\mathcal{F} = \mathcal{F}_{\mathrm{global}} \cup \mathcal{F}_{\mathrm{region}} \cup \mathcal{F}_{\mathrm{channel}}$ denote the complete feature set partitioned by spatial scale, and let $\mathcal{G} = \{\mathrm{global}, \mathrm{region}, \mathrm{channel}\}$ be the set of granularities. A selection procedure is said to be \textit{granularity-aware} if it (i) partitions the feature space by spatial scale a priori, (ii) enforces per-granularity selection budgets $\{K_g\}_{g \in \mathcal{G}}$ rather than a single global budget, and (iii) evaluates contributions separately per granularity. Equivalently, the selection problem is formulated as:
\begin{equation}\label{eq:1}
\max_{\{S_g\}_{g \in \mathcal{G}}} \sum_{g \in \mathcal{G}} \text{Score}(S_g, y), \quad \text{s.t. } |S_g| \le K_g,\; S_g \subseteq \mathcal{F}_g,\; \forall g \in \mathcal{G},
\end{equation}
where $\text{Score}(S_g, y)$ measures the predictive utility of feature subset $S_g$ with respect to labels $y$. This constrained formulation explicitly prevents high-dimensional granularities from dominating the selected representation and ensures that each spatial scale contributes within its allocated budget.

Let each input EEG epoch be denoted as $\mathbf{X} \in \mathbb{R}^{C \times T}$, where $C$ is the number of channels and $T$ is the number of time samples. We extracted a multi-scale set of EEG features across three spatial granularities—global, region, and channel—to capture both linear and nonlinear characteristics of neural activity. A summary of the extracted feature families is provided in Table~\ref{tab:feature_summary}. Under the default configuration ($C=129$, $B_{\mathrm{freq}}=5$, $R=5$ brain regions, $P=15$ regional pairs, $M=2$ power metrics, $Q=3$ epoch-level summary statistics, and $K_{\mathrm{ms}}=4$ microstates), this yields 28 global, 400 regional, and 3096 channel-level features, totaling 3524 conceptual features. All features were aggregated across epochs using {three} statistics (mean, standard deviation, {and} median) and $z$-score normalized. EEG preprocessing included bandpass filtering (1--45~Hz), common average referencing, bad channel detection with spline interpolation, and epoch rejection. 
All preprocessing steps were applied independently to each subject's recording. Common average referencing and bad-channel interpolation were restricted to within-subject channel information without cross-subject aggregation, so that each recording remained an independent unit throughout preprocessing.

Log transformation was applied to power values, and a regularization term ($\epsilon=10^{-15}$) ensured numerical stability for power ratios.

For \textbf{global microstate features}, we identified EEG microstates using $K$-means clustering on global field power peaks~\cite{Michel2018,Koenig2002}. For each of the $K_{\mathrm{ms}}=4$ canonical microstates, we computed coverage, occurrence, and mean duration ($3K_{\mathrm{ms}}=12$ features), along with the microstate transition probability matrix ($K_{\mathrm{ms}}^2=16$ features) characterizing temporal segmentation and transition patterns~\cite{Lehmann1987}.

For \textbf{frequency-domain features}, power spectral density (PSD) was estimated using Welch's method~\cite{Welch1967} with a 2-second Hanning window and 50\% overlap. For each channel and frequency band, we computed log-absolute power and relative power (each yielding $CB=645$ features); region-level features were obtained by averaging across channels within each of $R=5$ brain regions ($2RB=50$ features). At the channel level, we additionally computed differential entropy ($CB=645$ features), log-absolute band-power ratios for $(\theta/\beta)$, $(\alpha/\beta)$, and $(\theta/\alpha)$ ($3C=387$ features), and aperiodic parameters by fitting the spectral power spectrum as $\log(\mathrm{PSD}(f)) = \omega_{ch} - \kappa_{ch}\log(f) + \eta(f)$, where $\kappa_{ch}$ is the aperiodic exponent and $\omega_{ch}$ is the offset ($2C=258$ features)~\cite{He2010,Donoghue2020}. 

For \textbf{regional asymmetry and connectivity}, we quantified inter-hemispheric asymmetry for each power metric ($2RB=50$ features) and estimated inter-regional functional connectivity using coherence and phase-locking value (PLV) across regional pairs. For connectivity, we used six regions (including the auricular-mastoid reference), yielding $P=\binom{6}{2}=15$ pairs, and two connectivity metrics (coherence and PLV), giving $2PB=150$ features. For long-duration paradigms, we extracted sliding-window mean, standard deviation, and slope of regional spectral power ($RBMQ=150$ features). To quantify signal complexity, we computed Hjorth parameters (activity, mobility, complexity) and spectral entropy ($4C=516$ features)~\cite{Hjorth1970,liang2015eeg}. Sample entropy was included as an optional feature ($C=129$ if enabled). All features were organized into three granularities: global-level (28 microstate features), region-level (400 features), and channel-level (3096 features). The complete feature taxonomy, including dimensionality breakdown per feature family, is summarized in Table~\ref{tab:feature_summary}.

\subsection*{Granularity-Aware Feature-Model Selection Engine}

As illustrated in Figure~\ref{fig:main}, the core of our framework comprises a feature-model selection engine that systematically identifies optimal feature representations and model families. Let $\mathcal{T}$ denote the set of EEG tasks (EC, EO, MW, SuS) and $\mathcal{L}$ denote the set of prediction labels ($p$-factor, attention, externalizing, internalizing). For each task--label pair $(t,l) \in \mathcal{T} \times \mathcal{L}$, we construct a feature matrix and corresponding binary label vector. The feature space is organized according to a reference feature manifest $\mathcal{F}$, which specifies the spatial granularity of each feature. Let $\mathcal{G} = \{\mathrm{global}, \mathrm{region}, \mathrm{channel}\}$ denote the three granularity levels, with granularity-specific budgets $\{K_g\}_{g \in \mathcal{G}}$ controlling the number of features retained per level. This engine operates in two stages, as detailed below.

\phantomsection
\subsubsection*{Stage 2A: Cross-task Model Family Screening (CMFS)}
\label{sec:cmfs}

Before performing feature selection and final pipeline optimization, we conducted a cross-task model family screening procedure to identify robust candidate classifiers. The goal was to reduce the model search space by selecting model families that consistently performed well across different EEG tasks and psychological labels, rather than determining the final model for any specific task--label pair. For each task--label pair $(t,l) \in \mathcal{T} \times \mathcal{L}$, we constructed a full-feature matrix $X_{t,l}$ (including all three granularities) and binary label vector $y_{t,l}$, with only basic preprocessing (missing-value handling, low-variance removal, and standardization) applied. A predefined candidate set $\mathcal{M}$ comprising linear models (Logistic Regression, ElasticNet), SVM (Linear SVM), tree-based ensembles (Random Forest), and boosting models (XGBoost) was evaluated, with the Dummy classifier included as a sanity-check baseline. For each model family, hyperparameters were tuned via inner cross-validation within each outer training fold using AUC as the optimization objective.

To summarize robustness, for each model family $m \in \mathcal{M}$, we computed its mean AUC, median AUC, average rank, number of first-place rankings, and number of top-two rankings across all $|\mathcal{T}| \times |\mathcal{L}|$ task--label combinations:
\begin{equation}
\text{MeanRank}(m) = \frac{1}{|\mathcal{T}| \cdot |\mathcal{L}|} \sum_{t \in \mathcal{T}} \sum_{l \in \mathcal{L}} \text{Rank}(m, t, l)
\end{equation}
where $\text{Rank}(m, t, l)$ denotes the rank of model family $m$ on task--label pair $(t,l)$ according to validation AUC. The average rank assessed consistency, while top-two frequency indicated whether a model remained competitive even when not the best for a specific task. 
This screening produced a reduced set of candidate model families $\mathcal{M}^{\ast} \subset \mathcal{M}$ for subsequent experiments. By selecting based on cross-task stability rather than single-task performance, the framework avoids over-specialization to any particular EEG condition or psychological label. CMFS is a preliminary screening step. Its aggregated ranking statistics are derived from nested CV results on the full dataset. The step selects {two} out of {five} coarse model families, with no selection of hyperparameters or features. Downstream training is performed de novo within each outer fold of the GSTS pipeline.

\begin{algorithm}[t]
\small

\caption{Granularity-Aware Stability Top-$K$ Selection (GSTS)}
\label{alg:granularity_feature_selection}
\KwInput{
Training feature matrix $X_{train}$;
training labels $y_{train}$;
feature manifest $\mathcal{F}$;
granularity set $\mathcal{G}=\{\mathrm{global},\mathrm{region},\mathrm{channel}\}$;
granularity-specific budgets $\{K_g\}_{g \in \mathcal{G}}$;
base scoring method $\phi$;
number of resampling iterations $B_{\mathrm{stab}}$.
}
\KwOutput{
Selected feature set $\mathcal{S}^{\ast}$.
}
Initialize final selected feature set $\mathcal{S}^{\ast} \leftarrow \emptyset$\;
\ForEach{granularity $g \in \mathcal{G}$}{
    Identify feature subset
    $\mathcal{F}_g = \{f \in \mathcal{F}: \mathrm{granularity}(f)=g\}$\;
    Initialize selection count $c_f \leftarrow 0$ for each feature $f \in \mathcal{F}_g$\;
    \For{$b \leftarrow 1$ \KwTo $B_{\mathrm{stab}}$}{
        Draw a stratified subsample $\mathcal{D}_b$ from $(X_{train}, y_{train})$\;
        Extract $X_b^{(g)}$ using only features in $\mathcal{F}_g$\;
        Apply preprocessing using only $\mathcal{D}_b$\;
        Compute feature importance score $s_f$ for each feature $f \in \mathcal{F}_g$ using scoring method $\phi$\; 
        {\color{blue}Select the top-$K_g$ features in $\mathcal{F}_g$ according to $s_f$\;}  
        Increase $c_f$ by one for each selected feature\;
    }
    \ForEach{feature $f \in \mathcal{F}_g$}{
        Compute stability score $\pi_f = c_f / B_{\mathrm{stab}}$\;
    }
    Select $\mathcal{S}_g$, the top-$K_g$ features in $\mathcal{F}_g$ according to stability score $\pi_f$\;
    Update $\mathcal{S}^{\ast} \leftarrow \mathcal{S}^{\ast} \cup \mathcal{S}_g$\;
}
\Return $\mathcal{S}^{\ast}$\;
\end{algorithm}

\subsubsection*{Stage 2B: Granularity-Aware Stability Top-$K$ Feature Selection (GSTS)}

The detailed procedure of the proposed Granularity-Aware Stability Top-$K$ Selection (GSTS) framework is summarized in Algorithm~\ref{alg:granularity_feature_selection}. GSTS is designed to address a structural imbalance that naturally arises in multi-scale EEG feature spaces. The extracted feature set contains descriptors from three spatial granularities: global, regional, and channel levels. These granularities differ not only in dimensionality but also in neurophysiological meaning. Global features summarize whole-brain temporal dynamics, regional features capture anatomically aggregated functional patterns, and channel-level features preserve fine-grained topographical information. However, because channel-level descriptors are far more numerous than global or regional descriptors, a conventional flat selection strategy over the pooled feature space can introduce a candidate-pool bias: high-dimensional feature groups are more likely to occupy the selected subset simply because they contribute more candidates, even when lower-dimensional groups contain stable and complementary information.

To mitigate this bias, GSTS reformulates feature selection as a structured, granularity-constrained stability selection problem rather than a single flat ranking problem. Specifically, the complete feature space $\mathcal{F}$ is first partitioned according to the feature manifest into three non-overlapping subsets,
$\mathcal{F}_{\mathrm{global}}$, $\mathcal{F}_{\mathrm{region}}$, and $\mathcal{F}_{\mathrm{channel}}$. Instead of allowing all features to compete in one shared pool, GSTS performs competition within each granularity independently. Each granularity $g \in \mathcal{G}$ is assigned an independent feature budget $K_g$, so that the final representation is constrained to preserve information from multiple spatial scales. This design transforms the original imbalanced $K$-of-$N$ selection problem into a set of granularity-specific $K_g$-of-$N_g$ selection problems, where $N_g = |\mathcal{F}_g|$. As a result, each spatial scale contributes according to its within-granularity relevance rather than its raw dimensionality.

Within each granularity, GSTS further introduces resampling-based stability regularization. Given the training data only, GSTS repeatedly draws $B_{\mathrm{stab}}$ stratified subsamples and computes feature relevance scores within each granularity using a base scoring function $\phi$, such as the ANOVA F-test. In each resampling iteration $b$, the top-$K_g$ features from granularity $g$ are retained, and the selection count $c_f$ of each feature is updated. After $B_{\mathrm{stab}}$ iterations, every feature $f \in \mathcal{F}_g$ receives a stability score:
\begin{equation}
\pi_f = \frac{c_f}{B_{\mathrm{stab}}}, \quad
c_f = \sum_{b=1}^{B_{\mathrm{stab}}} \mathbb{I}(f \in \mathcal{S}_b^{(g)}),
\end{equation}
where $\mathcal{S}_b^{(g)}$ denotes the subset of top-$K_g$ features selected from granularity $g$ in the $b$-th resampling iteration.

For each granularity, GSTS selects the final subset $\mathcal{S}_g$ by retaining the $K_g$ features with the highest stability scores. The final selected representation is then obtained by merging the stable subsets from all granularities:
\begin{equation}
\mathcal{S}^{\ast} =
\mathcal{S}_{\mathrm{global}}
\cup
\mathcal{S}_{\mathrm{region}}
\cup
\mathcal{S}_{\mathrm{channel}}.
\end{equation}

This procedure provides three forms of regularization simultaneously: dimensionality reduction through top-$K_g$ selection, stability regularization through repeated subsampling, and granularity regularization through explicit multi-scale budget constraints. Therefore, GSTS does not merely reduce the number of EEG features; it preserves a balanced and interpretable multi-granularity representation while preventing high-dimensional channel-level descriptors from overwhelming lower-dimensional but neurophysiologically meaningful global and regional markers.

All GSTS operations are conducted strictly within the training folds of the nested cross-validation pipeline. Feature relevance scores, stability estimates, and final selected subsets are computed without access to the held-out test fold, and the learned feature subset is then applied unchanged to the corresponding test data. This fold-contained implementation ensures that the reported performance reflects genuine generalization rather than information leakage from feature selection. Unlike conventional feature selection methods that impose sparsity on the pooled feature space, GSTS imposes sparsity under granularity-aware structural constraints, thereby aligning feature reduction with the hierarchical organization of EEG descriptors.

\subsection*{Fold-Contained Supervised Evaluation Protocol}

To rigorously evaluate the proposed framework and prevent data leakage, we adopted a nested cross-validation (CV) protocol with leakage-free design, as illustrated in Figure~\ref{fig:main}. The outer loop consists of $K_{\mathrm{cv}}$-fold stratified cross-validation (default $K_{\mathrm{cv}}=5$), where each fold reserves one subject group as the test set while the remaining groups are used for training. All feature selection (GSTS), hyperparameter tuning, and model fitting are performed exclusively within each training fold using inner cross-validation (also 5-fold), and the held-out test fold is used only for final evaluation. This design ensures that the test data never participate in any decision-making process—including preprocessing, feature selection, or hyperparameter optimization—thereby providing unbiased estimates of generalization performance. {All EEG preprocessing operations (filtering, re-referencing, bad-channel interpolation) are applied independently to each subject's recording without cross-subject information exchange and are performed once before cross-validation splitting.}

For model selection, we used the five candidate model families (plus the Dummy classifier) screened via CMFS as described in \hyperref[sec:cmfs]{Stage 2A}. Hyperparameters for each model were optimized via grid search within the inner CV loop using AUC as the objective. The top-performing model families $\mathcal{M}^{\ast}$ were selected based on their cross-task stability across all task--label combinations, measured by mean rank, mean AUC, and top-two frequency. For feature selection, we employed GSTS with $B_{\mathrm{stab}}=100$ and ANOVA F-test as the base scoring method $\phi$. The feature budgets were set to $K_{\mathrm{global}}=10$, $K_{\mathrm{region}}=50$, and $K_{\mathrm{channel}}=100$ by default. All selection operations were performed using only the training data, and the selected subset $\mathcal{S}^{\ast}$ was applied to the test fold without refitting.

After completing all $K_{\mathrm{cv}}$ outer folds, performance metrics (AUC, balanced accuracy, and macro F1-score) are aggregated and reported as mean $\pm$ standard deviation. {To assess statistical significance, we employed two complementary procedures. For the full model-performance table (Table~\ref{tab:full_performance}), paired $t$-tests compared each model's fold-wise AUCs against the Dummy classifier within each task--label pair. For the ROC summary figure (Figure~\ref{fig:gsts_roc}), permutation tests with 1,000 random label shuffles were performed for each task--label pair, and the resulting $p$-values were corrected for multiple comparisons across all panels using false discovery rate (FDR) control.} This yields the optimized multi-granularity pipeline with interpretable outputs, including 95\% confidence intervals, stable EEG markers (robust features), and brain-region and frequency-band contributions.

\section*{Results and Discussion}

\begin{sidewaystable}[htbp]

\centering
\caption{Model performance on \textbf{resting-state and task-state EEG data} using the \textbf{full F-all feature set} (AUC / Accuracy / Macro F1) with nested 5-fold cross-validation [mean $\pm$ std]. Values are reported as mean $\pm$ standard deviation across 5 outer folds. For each paradigm and psychopathology label, the best-performing model was selected according to mean cross-validated AUC. Significance markers indicate paired t-test results (vs.\ Dummy Classifier) for AUC only: ${}^{*}p<0.05$, ${}^{**}p<0.01$. Best results per metric and per paradigm are shown in bold and colored.}

\begin{threeparttable}
\renewcommand{\arraystretch}{1.25}
  \resizebox{1\linewidth}{!}{
\begin{tabular}{lcccccccccccc}
\toprule
\multirow{2}{*}{\textbf{Model}} & \multicolumn{3}{c}{\textbf{P-factor}} & \multicolumn{3}{c}{\textbf{Attention}} & \multicolumn{3}{c}{\textbf{Externalizing}} & \multicolumn{3}{c}{\textbf{Internalizing}} \\
\cmidrule(lr){2-4} \cmidrule(lr){5-7} \cmidrule(lr){8-10} \cmidrule(lr){11-13}
 & \textbf{AUC} & \textbf{Accuracy} & \textbf{Macro F1} & \textbf{AUC} & \textbf{Accuracy} & \textbf{Macro F1} & \textbf{AUC} & \textbf{Accuracy} & \textbf{Macro F1} & \textbf{AUC} & \textbf{Accuracy} & \textbf{Macro F1} \\
\midrule
\multicolumn{13}{l}{\textbf{\color{blue}Resting-State paradigm 1 (Eye-closed EEG paradigm):}} \\
Dummy Classifier & $0.517 \pm 0.026$ & $0.517 \pm 0.026$ & $0.517 \pm 0.026$ & $0.484 \pm 0.054$ & $0.484 \pm 0.054$ & $0.483 \pm 0.054$ & $0.500 \pm 0.035$ & $0.500 \pm 0.035$ & $0.499 \pm 0.035$ & $0.483 \pm 0.018$ & $0.483 \pm 0.018$ & $0.482 \pm 0.018$ \\
Logistic Regression L2 & $0.501 \pm 0.022$ & $0.503 \pm 0.008$ & $0.502 \pm 0.008$ & $0.559 \pm 0.026$ & $0.534 \pm 0.023$ & $0.533 \pm 0.023$ & $0.547 \pm 0.019^{*}$ & $0.542 \pm 0.032$ & $0.542 \pm 0.032$ & $0.618 \pm 0.012^{**}$ & $0.593 \pm 0.013^{**}$ & $0.592 \pm 0.012^{**}$ \\
Logistic Reg. ElasticNet & $0.524 \pm 0.021$ & $0.526 \pm 0.027$ & $0.525 \pm 0.026$ & $0.550 \pm 0.026$ & $0.535 \pm 0.023$ & $0.533 \pm 0.022$ & $0.538 \pm 0.020$ & $0.526 \pm 0.024$ & $0.526 \pm 0.023$ & $0.581 \pm 0.020^{**}$ & $0.559 \pm 0.024^{*}$ & $0.557 \pm 0.026^{*}$ \\
Linear SVM & $0.493 \pm 0.026$ & $0.500 \pm 0.028$ & $0.498 \pm 0.028$ & $0.550 \pm 0.031$ & $0.535 \pm 0.016$ & $0.535 \pm 0.016$ & $0.546 \pm 0.020$ & $0.537 \pm 0.024$ & $0.535 \pm 0.025$ & $0.598 \pm 0.011^{**}$ & $0.570 \pm 0.012^{**}$ & $0.568 \pm 0.013^{**}$ \\
Random Forest & $0.534 \pm 0.034$ & $0.525 \pm 0.028$ & $0.524 \pm 0.028$ & $0.553 \pm 0.049$ & $0.542 \pm 0.037$ & $0.541 \pm 0.037$ & \cellcolor{orange!20}$\mathbf{0.605 \pm 0.028^{**}}$ & \cellcolor{orange!20}$\mathbf{0.582 \pm 0.028^{**}}$ & \cellcolor{orange!20}$\mathbf{0.582 \pm 0.028^{**}}$ & \cellcolor{blue!20}$\mathbf{0.655 \pm 0.017^{**}}$ & \cellcolor{blue!20}$\mathbf{0.626 \pm 0.029^{**}}$ & \cellcolor{blue!20}$\mathbf{0.626 \pm 0.029^{**}}$ \\
XGBoost & \cellcolor{red!20}$\mathbf{0.549 \pm 0.004}$ & \cellcolor{red!20}$\mathbf{0.528 \pm 0.011}$ & \cellcolor{red!20}$\mathbf{0.527 \pm 0.011}$ & \cellcolor{green!20}$\mathbf{0.576 \pm 0.035}$ & \cellcolor{green!20}$\mathbf{0.549 \pm 0.030}$ & \cellcolor{green!20}$\mathbf{0.548 \pm 0.031}$ & $0.597 \pm 0.025^{**}$ & $0.579 \pm 0.024^{**}$ & $0.579 \pm 0.024^{**}$ & $0.651 \pm 0.024^{**}$ & $0.609 \pm 0.006^{**}$ & $0.609 \pm 0.006^{**}$ \\
\midrule
\multicolumn{13}{l}{\textbf{\color{blue}Resting-State paradigm 2 (Eye-open EEG paradigm):}} \\
Dummy Classifier & $0.510 \pm 0.018$ & $0.510 \pm 0.018$ & $0.510 \pm 0.018$ & $0.500 \pm 0.051$ & $0.500 \pm 0.051$ & $0.500 \pm 0.052$ & $0.509 \pm 0.025$ & $0.509 \pm 0.025$ & $0.508 \pm 0.025$ & $0.499 \pm 0.025$ & $0.499 \pm 0.025$ & $0.499 \pm 0.025$ \\
Logistic Regression L2 & $0.516 \pm 0.018$ & $0.510 \pm 0.026$ & $0.510 \pm 0.026$ & $0.552 \pm 0.035$ & $0.549 \pm 0.020$ & $0.549 \pm 0.020$ & $0.559 \pm 0.020^{*}$ & $0.564 \pm 0.020^{*}$ & $0.564 \pm 0.020^{*}$ & $0.598 \pm 0.015^{**}$ & $0.581 \pm 0.013^{**}$ & $0.580 \pm 0.013^{**}$ \\
Logistic Reg. ElasticNet & $0.509 \pm 0.018$ & $0.514 \pm 0.023$ & $0.513 \pm 0.023$ & $0.548 \pm 0.035$ & $0.540 \pm 0.029$ & $0.537 \pm 0.028$ & $0.566 \pm 0.029^{*}$ & $0.552 \pm 0.031$ & $0.550 \pm 0.034$ & $0.566 \pm 0.032$ & $0.561 \pm 0.035$ & $0.560 \pm 0.036$ \\
Linear SVM & $0.517 \pm 0.013$ & $0.499 \pm 0.029$ & $0.498 \pm 0.029$ & $0.553 \pm 0.036$ & \cellcolor{green!20}$\mathbf{0.551 \pm 0.022}$ & \cellcolor{green!20}$\mathbf{0.551 \pm 0.023}$ & $0.541 \pm 0.021^{*}$ & $0.526 \pm 0.014$ & $0.524 \pm 0.013$ & $0.576 \pm 0.021^{*}$ & $0.567 \pm 0.009^{**}$ & $0.566 \pm 0.009^{**}$ \\
Random Forest & $0.516 \pm 0.028$ & $0.508 \pm 0.018$ & $0.507 \pm 0.018$ & \cellcolor{green!20}$\mathbf{0.553 \pm 0.040}$ & $0.550 \pm 0.031$ & $0.549 \pm 0.031$ & \cellcolor{orange!20}$\mathbf{0.623 \pm 0.040^{**}}$ & \cellcolor{orange!20}$\mathbf{0.605 \pm 0.046^{*}}$ & \cellcolor{orange!20}$\mathbf{0.605 \pm 0.046^{*}}$ & $0.643 \pm 0.025^{**}$ & $0.608 \pm 0.020^{**}$ & $0.607 \pm 0.019^{**}$ \\
XGBoost & \cellcolor{red!20}$\mathbf{0.527 \pm 0.028}$ & \cellcolor{red!20}$\mathbf{0.526 \pm 0.018}$ & \cellcolor{red!20}$\mathbf{0.526 \pm 0.018}$ & $0.522 \pm 0.042$ & $0.521 \pm 0.030$ & $0.520 \pm 0.030$ & $0.612 \pm 0.053^{*}$ & $0.575 \pm 0.038^{*}$ & $0.574 \pm 0.038$ & \cellcolor{blue!20}$\mathbf{0.663 \pm 0.024^{**}}$ & \cellcolor{blue!20}$\mathbf{0.623 \pm 0.024^{**}}$ & \cellcolor{blue!20}$\mathbf{0.623 \pm 0.024^{**}}$ \\
\midrule
\multicolumn{13}{l}{\textbf{\color{blue}Task-State paradigm 1 (Movie-Watching EEG paradigm):}} \\
Dummy Classifier & $0.505 \pm 0.028$ & $0.505 \pm 0.028$ & $0.505 \pm 0.028$ & $0.489 \pm 0.026$ & $0.489 \pm 0.026$ & $0.489 \pm 0.026$ & $0.483 \pm 0.016$ & $0.483 \pm 0.016$ & $0.483 \pm 0.016$ & $0.483 \pm 0.025$ & $0.483 \pm 0.025$ & $0.483 \pm 0.025$ \\
Logistic Regression L2 & $0.518 \pm 0.036$ & $0.510 \pm 0.031$ & $0.508 \pm 0.030$ & $0.567 \pm 0.023^{**}$ & $0.547 \pm 0.013^{**}$ & \cellcolor{green!20}$\mathbf{0.547 \pm 0.014^{**}}$ & $0.603 \pm 0.030^{**}$ & $0.575 \pm 0.035^{**}$ & $0.574 \pm 0.034^{**}$ & $0.571 \pm 0.054^{*}$ & $0.548 \pm 0.057$ & $0.548 \pm 0.057$ \\
Logistic Reg. ElasticNet & $0.507 \pm 0.039$ & $0.505 \pm 0.033$ & $0.499 \pm 0.039$ & $0.550 \pm 0.009^{**}$ & \cellcolor{green!20}$\mathbf{0.548 \pm 0.016^{**}}$ & $0.544 \pm 0.021^{**}$ & $0.526 \pm 0.037^{*}$ & $0.528 \pm 0.037^{*}$ & $0.526 \pm 0.037^{*}$ & $0.556 \pm 0.036^{*}$ & $0.546 \pm 0.029^{*}$ & $0.545 \pm 0.028^{*}$ \\
Linear SVM & $0.509 \pm 0.038$ & $0.510 \pm 0.027$ & $0.509 \pm 0.028$ & $0.550 \pm 0.033$ & $0.537 \pm 0.028$ & $0.536 \pm 0.029$ & $0.534 \pm 0.041$ & $0.531 \pm 0.019^{**}$ & $0.530 \pm 0.019^{**}$ & $0.545 \pm 0.051$ & $0.532 \pm 0.048$ & $0.530 \pm 0.049$ \\
Random Forest & $0.517 \pm 0.037$ & \cellcolor{red!20}$\mathbf{0.524 \pm 0.025}$ & \cellcolor{red!20}$\mathbf{0.524 \pm 0.026}$ & $0.539 \pm 0.028$ & $0.522 \pm 0.022$ & $0.521 \pm 0.021$ & \cellcolor{orange!20}$\mathbf{0.640 \pm 0.025^{**}}$ & \cellcolor{orange!20}$\mathbf{0.610 \pm 0.026^{**}}$ & \cellcolor{orange!20}$\mathbf{0.610 \pm 0.026^{**}}$ & \cellcolor{blue!20}$\mathbf{0.623 \pm 0.051^{**}}$ & \cellcolor{blue!20}$\mathbf{0.588 \pm 0.045^{**}}$ & \cellcolor{blue!20}$\mathbf{0.587 \pm 0.046^{**}}$ \\
XGBoost & \cellcolor{red!20}$\mathbf{0.529 \pm 0.029}$ & $0.520 \pm 0.032$ & $0.518 \pm 0.031$ & \cellcolor{green!20}$\mathbf{0.570 \pm 0.034^{**}}$ & $0.546 \pm 0.026^{*}$ & $0.546 \pm 0.026^{*}$ & $0.620 \pm 0.024^{**}$ & $0.594 \pm 0.016^{**}$ & $0.594 \pm 0.016^{**}$ & $0.620 \pm 0.036^{**}$ & $0.575 \pm 0.031^{**}$ & $0.574 \pm 0.031^{**}$ \\
\midrule
\multicolumn{13}{l}{\textbf{\color{blue}Task-State paradigm 2 (Surround Suppression EEG paradigm):}} \\
Dummy Classifier & $0.471 \pm 0.026$ & $0.471 \pm 0.026$ & $0.471 \pm 0.026$ & $0.506 \pm 0.038$ & $0.506 \pm 0.038$ & $0.506 \pm 0.038$ & $0.484 \pm 0.028$ & $0.484 \pm 0.028$ & $0.484 \pm 0.028$ & $0.490 \pm 0.049$ & $0.490 \pm 0.049$ & $0.490 \pm 0.049$ \\
Logistic Regression L2 & $0.509 \pm 0.046$ & $0.492 \pm 0.048$ & $0.491 \pm 0.049$ & $0.492 \pm 0.032$ & $0.501 \pm 0.034$ & $0.500 \pm 0.034$ & $0.532 \pm 0.027^{*}$ & $0.522 \pm 0.021$ & $0.522 \pm 0.021$ & $0.583 \pm 0.032$ & $0.551 \pm 0.026$ & $0.549 \pm 0.027$ \\
Logistic Reg. ElasticNet & $0.493 \pm 0.033$ & $0.483 \pm 0.028$ & $0.481 \pm 0.028$ & $0.510 \pm 0.045$ & $0.526 \pm 0.048$ & $0.523 \pm 0.048$ & $0.529 \pm 0.034$ & $0.521 \pm 0.027$ & $0.519 \pm 0.024$ & $0.551 \pm 0.031$ & $0.549 \pm 0.020$ & $0.546 \pm 0.021$ \\
Linear SVM & $0.505 \pm 0.049$ & $0.509 \pm 0.045$ & $0.506 \pm 0.045$ & $0.490 \pm 0.041$ & $0.491 \pm 0.026$ & $0.487 \pm 0.026$ & $0.502 \pm 0.020$ & $0.504 \pm 0.014$ & $0.501 \pm 0.010$ & $0.559 \pm 0.033$ & $0.537 \pm 0.031$ & $0.536 \pm 0.032$ \\
Random Forest & $0.541 \pm 0.035^{*}$ & \cellcolor{red!20}$\mathbf{0.529 \pm 0.026}$ & \cellcolor{red!20}$\mathbf{0.529 \pm 0.026}$ & $0.543 \pm 0.047$ & \cellcolor{green!20}$\mathbf{0.530 \pm 0.035}$ & \cellcolor{green!20}$\mathbf{0.528 \pm 0.036}$ & \cellcolor{orange!20}$\mathbf{0.602 \pm 0.051^{*}}$ & \cellcolor{orange!20}$\mathbf{0.573 \pm 0.032^{**}}$ & \cellcolor{orange!20}$\mathbf{0.572 \pm 0.034^{**}}$ & $0.633 \pm 0.039^{*}$ & $0.602 \pm 0.028^{*}$ & $0.602 \pm 0.028^{*}$ \\
XGBoost & \cellcolor{red!20}$\mathbf{0.544 \pm 0.016^{**}}$ & $0.520 \pm 0.015^{**}$ & $0.520 \pm 0.015^{**}$ & \cellcolor{green!20}$\mathbf{0.560 \pm 0.042}$ & $0.527 \pm 0.037$ & $0.526 \pm 0.037$ & $0.592 \pm 0.052^{*}$ & $0.564 \pm 0.052^{*}$ & $0.563 \pm 0.053^{*}$ & \cellcolor{blue!20}$\mathbf{0.645 \pm 0.036^{*}}$ & \cellcolor{blue!20}$\mathbf{0.609 \pm 0.018^{*}}$ & \cellcolor{blue!20}$\mathbf{0.609 \pm 0.017^{*}}$ \\

\bottomrule
\end{tabular}
}
\end{threeparttable}
\label{tab:full_performance}
\end{sidewaystable}

\begin{table}[t]
\centering
\caption{\textbf{Cross-task model family screening results.} Models were evaluated on the full feature set (F-all) across all 16 (task, label) combinations from 4 paradigms (EC, EO, Movie, SuS). For each combination, models were ranked by validation AUC. Mean rank and median AUC summarize cross-task stability; \#Best and \#Top2 indicate how often a model achieved the first or top-two distinct AUC ranks. Models with mean rank $\le 3$ and at least one Top2 occurrence were selected for subsequent experiments.}
\label{tab:model_screening}
\renewcommand{\arraystretch}{1.15}
\resizebox{0.8\linewidth}{!}{%
\begin{tabular}{lcccccc}
\toprule
Model Family & Mean AUC $\pm$ SD & Median AUC & Mean Rank $\downarrow$ & \#Best & \#Top2 & Decision \\
\midrule
\rowcolor{gray!15}
XGBoost      & $0.586 \pm 0.045$ & 0.584 & 1.62 & 9  & 15 & \textbf{Selected} \\
\rowcolor{gray!15}
Random Forest & $0.583 \pm 0.048$ & 0.577 & 2.00 & 7  & 12 & \textbf{Selected} \\
L2 Logistic  & $0.552 \pm 0.037$ & 0.555 & 3.12 & 0  & 3  & Reserve \\
ElasticNet   & $0.538 \pm 0.024$ & 0.543 & 4.38 & 0  & 0  & Reserve \\
Linear SVM   & $0.536 \pm 0.030$ & 0.543 & 4.19 & 0  & 2  & Reserve \\
Dummy        & $0.494 \pm 0.014$ & 0.496 & 5.69 & 0  & 0  & -- \\
\bottomrule
\end{tabular}
}
\end{table}

\subsection*{Analysis of experimental results} \label{sec:experimental_results}

{To establish a systematic baseline for EEG-based psychopathology prediction across diverse recording conditions and clinical dimensions, we first evaluated the predictive performance of six candidate models, ranging from linear classifiers to tree-based ensembles, using the full feature set across four EEG paradigms and four psychopathology dimensions (Table~\ref{tab:full_performance}).}
Tree-based ensembles (XGBoost and Random Forest) consistently outperformed linear models across all 16 task--label combinations, together accounting for every best-AUC position: XGBoost led in 9 of 16 cases (AUC range 0.522--0.663) and Random Forest in the remaining 7 (range 0.516--0.655). Linear models (L2 Logistic, ElasticNet, Linear SVM) produced consistently lower AUCs (range 0.490--0.618), confirming that EEG--psychopathology relationships are predominantly nonlinear. Across psychopathology dimensions, internalizing was the most predictable (best AUC per paradigm 0.623--0.663), followed by externalizing (0.602--0.640), while $p$-factor (0.527--0.549) and attention (0.553--0.576) showed lower predictability. Among paradigms, resting-state (EC and EO) and movie-watching produced comparable predictive signals, while surround suppression yielded the weakest overall performance. {{Paired $t$-tests against the Dummy classifier} confirmed that the AUCs of tree-based models were significantly above chance ( \(p < 0.05\) or \(p < 0.01\)) for most task--label pairs, whereas linear models frequently failed to reach significance, particularly for $p$-factor and attention dimensions.}

{This may be attributed to the fact that the block-design SSVEP paradigm in the SuS task introduces strong, periodic visual-evoked responses that could mask the more subtle, endogenous neural signals associated with trait-level psychopathology, making it less suitable for capturing individual differences compared to spontaneous resting-state or naturalistic viewing conditions.}

\subsection*{Model Screening Results} \label{sec:Model_Screening}

Given the consistent superiority of tree-based ensembles over linear models observed in the full-feature evaluation (Table~\ref{tab:full_performance}), we next conducted a formal cross-task model screening procedure to determine which specific model families should be retained for downstream experiments. Using cross-task stability—defined as the ability to maintain competitive performance across four EEG paradigms and four psychopathology dimensions—as the selection criterion, we ranked models by validation AUC for each of the 16 (task, label) combinations under nested 5-fold CV (Table~\ref{tab:model_screening}). 
XGBoost achieved the highest mean AUC (0.586) and lowest mean rank (1.62), followed closely by Random Forest (0.583, rank 2.00); XGBoost also delivered the most first-place wins (9/16) and the highest Top-2 frequency (15/16), with Random Forest ranking second in both metrics (7/16 first-place, 12/16 Top-2). Linear models (L2 Logistic, ElasticNet, Linear SVM) trailed substantially, with mean ranks ranging from 3.12 to 4.38 and zero first-place wins. {Applying a selection criterion of mean rank \(\le 3\) and at least one Top-2 occurrence, we retained XGBoost and Random Forest as an equal-weight ensemble (prediction-score level) for all downstream experiments; L2 Logistic (rank 3.12, 3 Top-2) was reserved as a baseline comparison, while ElasticNet (rank 4.38, 0 Top-2) and Linear SVM (rank 4.19, 2 Top-2) were excluded.}

\begin{table}[t]
\centering
\caption{\textbf{Ablation study of feature granularity on EEG-based psychopathology prediction.} The ensemble (XGBoost + Random Forest) was evaluated under 5-fold cross-validation on eyes-closed (EC) resting-state and movie-watching (MW) paradigms. Results are reported as AUC / Accuracy / Macro F1 (mean $\pm$ std). \#F counts conceptual features; each is aggregated via three statistics (mean, std, median), yielding 3× columns in the final model input matrix. Best results per target and paradigm are highlighted in color.
}
\label{tab:granularity_ablation_ensemble}
\begin{threeparttable}
\renewcommand{\arraystretch}{1.0}
\resizebox{1\linewidth}{!}{
\begin{tabular}{lccccccccccccc}
\toprule
\multirow{2}{*}{\bf Feature Granularity} 
& \multirow{2}{*}{\bf \#F}
& \multicolumn{3}{c}{\bf P-factor} 
& \multicolumn{3}{c}{\bf Attention} 
& \multicolumn{3}{c}{\bf Externalizing} 
& \multicolumn{3}{c}{\bf Internalizing} \\
\cmidrule(lr){3-5} 
\cmidrule(lr){6-8} 
\cmidrule(lr){9-11} 
\cmidrule(lr){12-14}
& 
& \textbf{AUC} & \textbf{Accuracy} & \textbf{Macro F1} 
& \textbf{AUC} & \textbf{Accuracy} & \textbf{Macro F1} 
& \textbf{AUC} & \textbf{Accuracy} & \textbf{Macro F1} 
& \textbf{AUC} & \textbf{Accuracy} & \textbf{Macro F1} \\
\midrule

\multicolumn{14}{l}{\textbf{\color{blue}Resting-State Paradigm 1 (Eye-closed EEG Paradigm):}} \\
  Global & 28 & $0.512 \pm 0.030$ & $0.508 \pm 0.027$ & $0.507 \pm 0.028$ & $0.505 \pm 0.033$ & $0.498 \pm 0.026$ & $0.497 \pm 0.025$ & $0.583 \pm 0.019$ & $0.556 \pm 0.018$ & $0.556 \pm 0.018$ & $0.582 \pm 0.030$ & $0.558 \pm 0.024$ & $0.558 \pm 0.024$ \\
  Region & 400 & $0.538 \pm 0.023$ & $0.522 \pm 0.019$ & $0.522 \pm 0.019$ & $0.581 \pm 0.037$ & \cellcolor{green!15}$0.561 \pm 0.023$ & \cellcolor{green!15}$0.561 \pm 0.023$ & $0.593 \pm 0.016$ & $0.560 \pm 0.020$ & $0.559 \pm 0.020$ & $0.644 \pm 0.019$ & $0.605 \pm 0.014$ & $0.605 \pm 0.014$ \\
  Channel & 3096 & $0.536 \pm 0.016$ & $0.521 \pm 0.010$ & $0.520 \pm 0.010$ & $0.583 \pm 0.043$ & $0.561 \pm 0.037$ & \cellcolor{green!15}$0.560 \pm 0.037$ & $0.598 \pm 0.029$ & $0.567 \pm 0.036$ & $0.566 \pm 0.036$ & $0.641 \pm 0.022$ & $0.606 \pm 0.037$ & $0.605 \pm 0.037$ \\
  Global + Region & 428 & $0.530 \pm 0.022$ & $0.526 \pm 0.010$ & $0.525 \pm 0.010$ & $0.569 \pm 0.041$ & $0.549 \pm 0.029$ & $0.548 \pm 0.029$ & $0.598 \pm 0.020$ & $0.572 \pm 0.023$ & $0.572 \pm 0.023$ & $0.646 \pm 0.024$ & $0.602 \pm 0.026$ & $0.602 \pm 0.026$ \\
  Global + Channel & 3124 & \cellcolor{red!15}$0.547 \pm 0.015$ & \cellcolor{red!15}$0.538 \pm 0.018$ & \cellcolor{red!15}$0.538 \pm 0.018$ & $0.569 \pm 0.038$ & $0.545 \pm 0.028$ & $0.544 \pm 0.028$ & $0.605 \pm 0.029$ & $0.575 \pm 0.021$ & $0.575 \pm 0.021$ & $0.645 \pm 0.023$ & $0.611 \pm 0.025$ & $0.610 \pm 0.025$ \\
  Region + Channel & 3496 & $0.520 \pm 0.009$ & $0.503 \pm 0.016$ & $0.502 \pm 0.016$ & \cellcolor{green!15}$0.584 \pm 0.043$ & $0.556 \pm 0.035$ & $0.556 \pm 0.035$ & \cellcolor{orange!15}$0.609 \pm 0.033$ & \cellcolor{orange!15}$0.583 \pm 0.032$ & \cellcolor{orange!15}$0.583 \pm 0.032$ & \cellcolor{blue!15}$0.660 \pm 0.022$ & \cellcolor{blue!15}$0.620 \pm 0.025$ & \cellcolor{blue!15}$0.620 \pm 0.025$ \\
  Global + Region + Channel & 3524 & $0.546 \pm 0.007$ & $0.534 \pm 0.019$ & $0.533 \pm 0.019$ & $0.570 \pm 0.047$ & $0.548 \pm 0.025$ & $0.548 \pm 0.025$ & $0.601 \pm 0.028$ & $0.562 \pm 0.031$ & $0.561 \pm 0.032$ & $0.659 \pm 0.027$ & $0.614 \pm 0.028$ & $0.613 \pm 0.028$ \\

\midrule

\multicolumn{14}{l}{\textbf{\color{blue}Task-State Paradigm 1 (Movie-Watching EEG Paradigm):}} \\

  Global & 28 & $0.518 \pm 0.025$ & $0.521 \pm 0.026$ & $0.519 \pm 0.025$ & $0.516 \pm 0.034$ & $0.521 \pm 0.025$ & $0.521 \pm 0.025$ & $0.593 \pm 0.014$ & $0.568 \pm 0.013$ & $0.568 \pm 0.013$ & $0.548 \pm 0.028$ & $0.529 \pm 0.025$ & $0.528 \pm 0.025$ \\
  Region & 400 & $0.541 \pm 0.029$ & \cellcolor{red!15}$0.537 \pm 0.022$ & \cellcolor{red!15}$0.536 \pm 0.021$ & $0.570 \pm 0.038$ & $0.566 \pm 0.018$ & $0.566 \pm 0.018$ & $0.622 \pm 0.029$ & $0.602 \pm 0.023$ & $0.601 \pm 0.023$ & $0.600 \pm 0.037$ & $0.580 \pm 0.029$ & $0.580 \pm 0.030$ \\
  Channel & 3096 & $0.496 \pm 0.029$ & $0.488 \pm 0.025$ & $0.487 \pm 0.025$ & $0.558 \pm 0.041$ & $0.542 \pm 0.037$ & $0.541 \pm 0.037$ & $0.623 \pm 0.017$ & \cellcolor{orange!15}$0.606 \pm 0.029$ & \cellcolor{orange!15}$0.605 \pm 0.029$ & $0.634 \pm 0.046$ & $0.592 \pm 0.028$ & $0.591 \pm 0.029$ \\
  Global + Region & 428 & \cellcolor{red!15}$0.546 \pm 0.034$ & $0.529 \pm 0.025$ & $0.528 \pm 0.025$ & \cellcolor{green!15}$0.586 \pm 0.024$ & \cellcolor{green!15}$0.567 \pm 0.017$ & \cellcolor{green!15}$0.567 \pm 0.018$ & $0.618 \pm 0.031$ & $0.592 \pm 0.039$ & $0.591 \pm 0.039$ & $0.613 \pm 0.034$ & $0.580 \pm 0.029$ & $0.580 \pm 0.030$ \\
  Global + Channel & 3124 & $0.501 \pm 0.017$ & $0.494 \pm 0.020$ & $0.493 \pm 0.019$ & $0.560 \pm 0.026$ & $0.550 \pm 0.023$ & $0.550 \pm 0.023$ & $0.617 \pm 0.022$ & $0.594 \pm 0.019$ & $0.594 \pm 0.019$ & $0.629 \pm 0.048$ & $0.605 \pm 0.033$ & $0.604 \pm 0.033$ \\
  Region + Channel & 3496 & $0.514 \pm 0.032$ & $0.515 \pm 0.028$ & $0.513 \pm 0.028$ & $0.568 \pm 0.045$ & $0.532 \pm 0.035$ & $0.532 \pm 0.035$ & \cellcolor{orange!15}$0.634 \pm 0.021$ & $0.605 \pm 0.034$ & $0.604 \pm 0.034$ & \cellcolor{blue!15}$0.643 \pm 0.053$ & \cellcolor{blue!15}$0.612 \pm 0.035$ & \cellcolor{blue!15}$0.612 \pm 0.035$ \\
  Global + Region + Channel & 3524 & $0.522 \pm 0.019$ & $0.527 \pm 0.025$ & $0.526 \pm 0.025$ & $0.570 \pm 0.038$ & $0.541 \pm 0.024$ & $0.541 \pm 0.024$ & $0.627 \pm 0.014$ & $0.591 \pm 0.017$ & $0.590 \pm 0.017$ & $0.634 \pm 0.044$ & $0.599 \pm 0.026$ & $0.599 \pm 0.026$ \\

\bottomrule
\end{tabular}
}

\end{threeparttable}
\end{table}

\subsection*{Feature Granularity Analysis} \label{sec:Feature_Granularity}

To assess the contribution of different spatial granularities to psychopathology prediction, we conducted an ablation study comparing seven feature configurations,  global-level (28 conceptual features), region-level (400 conceptual features), channel-level (3,096 conceptual features), and their pairwise and full combinations, using the selected ensemble (XGBoost and Random Forest) on both eyes-closed resting-state and movie-watching EEG datasets. 
From Table~\ref{tab:granularity_ablation_ensemble}, the \textbf{Region+Channel combination} consistently achieved the best overall performance across most dimensions and paradigms, yielding the highest AUCs for internalizing in both EC (0.660) and MW (0.643), externalizing in EC (0.609), and attention in EC (0.584). {This suggests that region-level and channel-level features provide complementary information: regional aggregation captures anatomically organized functional patterns, while channel-level details preserve fine-grained topographical nuances that jointly enhance predictive power.} 
Interestingly, \textbf{adding global microstate features} to Region+Channel (i.e., the full G+R+C set) did not further improve performance and, in several cases, slightly underperformed the Region+Channel combination (e.g., EC internalizing: 0.659 vs. 0.660; MW externalizing: 0.627 vs. 0.634). {This suggests that global microstate features capture whole-brain temporal dynamics that are qualitatively different from region- and channel-level spectral features; when combined, their information may overlap with what is already represented in the spatially resolved features, yielding marginal net gain rather than additive improvement. Rather than being uninformative, global features may serve a complementary role that is most valuable when region- and channel-level information is not yet present, but offers limited additional benefit once finer spatial details are accounted for.}

\begin{figure}[t]
\centering
\includegraphics[width=1\linewidth]{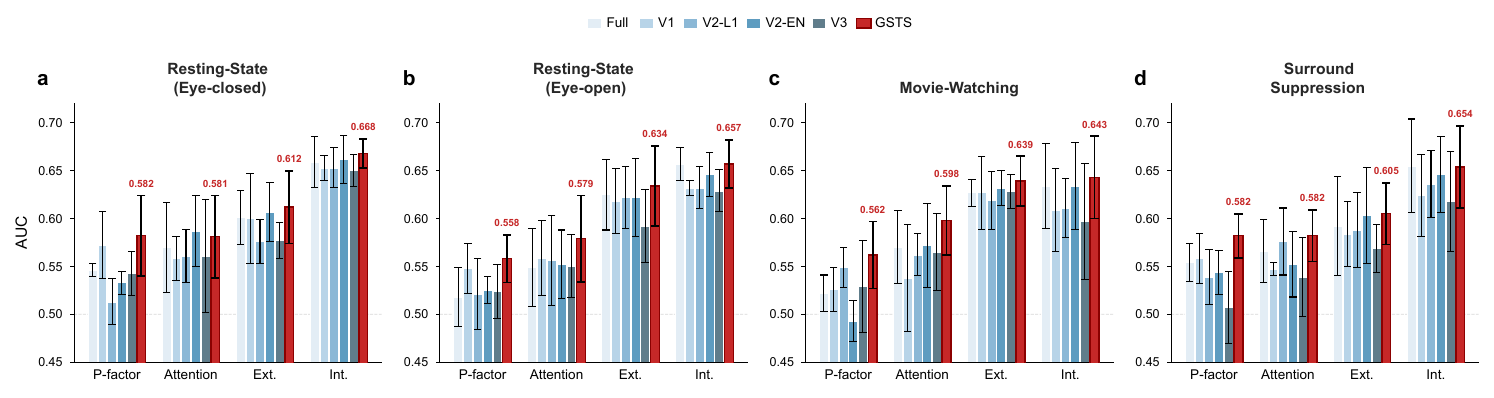}
\caption{\textbf{Per-paradigm comparison of feature selection strategies across four EEG paradigms.} (a) EC, (b) EO, (c) MW, and (d) SuS: AUC values (mean $\pm$ std) for the full-feature baseline, comparison variants, and the proposed GSTS framework across four psychopathology dimensions.}
\label{fig:ec_mw_comparison}
\end{figure}

\begin{table}[t]
\centering
\caption{\textbf{Comparison of feature selection strategies across four EEG paradigms (EC, EO, MW, SuS).}
All strategies are described in \hyperref[sec:Feature_Select_Strategy]{Feature Selection Strategy Analysis}.
Full denotes the no-selection baseline. V1 denotes flat stability selection, V2-L1 and V2-EN denote embedded L1 and ElasticNet selection, respectively, and V3 denotes PCA-based dimensionality reduction. GSTS denotes the proposed granularity-aware stability top-$K$ selection framework.
Adjusted stability ($J_{\mathrm{adj}}$)~\cite{liu2018survey} corrects for chance-level overlap expected from random selection:
$J_{\mathrm{adj}} = (J - J_{\mathrm{exp}}) / (1 - J_{\mathrm{exp}})$, where $J_{\mathrm{exp}} = k/(2N - k)$,
with $k$ being the number of selected features and $N = 10{,}572$ the total number of candidate features.
\#Top2 indicates how many task--label combinations a strategy ranked within the top two.
\textbf{Bold} indicates the best value among comparison strategies.
}
\label{tab:feature_select_aggregate}
\begin{threeparttable}
\renewcommand{\arraystretch}{1.15}
\resizebox{0.92\linewidth}{!}{%
\begin{tabular}{lccccc}
\toprule
\textbf{Strategy Variants} & $\mathbf{k}$ & \textbf{Adjusted Stability} $\uparrow$ & \textbf{Mean AUC} $\uparrow$ & \textbf{Mean Rank} $\downarrow$ & \textbf{\#Top2} $\uparrow$ \\
\midrule
Full: No selection & 10{,}572 & -- & $0.5903 \pm 0.0463$ & 3.31 & 3 \\
V1: Flat stability Top-200 & 200 & 0.359 & $0.5816 \pm 0.0403$ & 4.81 & 0 \\
V2-L1: Embedded L1 selection & 318 & 0.207 & $0.5821 \pm 0.0413$ & 4.56 & 2 \\
V2-EN: Embedded ElasticNet selection & 8{,}710 & 0.100 & $0.5882 \pm 0.0492$ & 3.31 & 7 \\
V3: PCA-based reduction & 200 & -- & $0.5593 \pm 0.0422$ & 6.56 & 0 \\
\textbf{GSTS: Granularity-aware selection (ours)} & 1{,}121 & $\mathbf{0.497}$ & $\mathbf{0.6082 \pm 0.0349}$ & \textbf{1.19} & \textbf{16} \\
GSTS-compact: Budget-constrained GSTS & 226 & 0.419 & $0.5901 \pm 0.0324$ & 4.00 & -- \\
\bottomrule
\end{tabular}
}

\end{threeparttable}
\end{table}

\phantomsection 
\subsection*{Feature Selection Strategy Analysis} 
\label{sec:Feature_Select_Strategy}

To evaluate whether the proposed granularity-aware selection strategy improves EEG-based psychopathology prediction, we compared it with five representative alternatives, including the full-feature baseline, flat stability selection, embedded sparse selection, ElasticNet selection, and PCA-based dimensionality reduction. 
The proposed method (Algorithm~\ref{alg:granularity_feature_selection}) is referred to as GSTS throughout this section.

The other methods are denoted as comparison variants: Full denotes the no-selection baseline, V1 denotes flat stability selection, V2-L1 and V2-EN denote embedded L1 and ElasticNet selection, respectively, and V3 denotes PCA-based reduction. 
Figure~\ref{fig:ec_mw_comparison} first presents the per-paradigm comparison across the four EEG conditions. Overall, GSTS achieved consistently strong AUC performance across EC, EO, MW, and SuS, whereas the competing variants showed more pronounced paradigm-dependent fluctuations. V2-EN obtained relatively competitive AUC values in some conditions, but it retained most of the original features, indicating limited dimensionality reduction. In contrast, V1 and V2-L1 produced compact feature subsets but yielded lower AUCs, suggesting that flat or globally sparse selection may remove weak but complementary EEG markers. V3 showed the lowest overall predictive performance, consistent with the fact that PCA maximizes unsupervised variance rather than label-discriminative information. These per-paradigm results indicate that GSTS provides a more favorable balance between predictive accuracy and feature reduction across heterogeneous EEG paradigms.

Table~\ref{tab:feature_select_aggregate} further summarizes the results across all four paradigms and all task--label combinations. On the primary performance metric, mean AUC, GSTS achieved the best overall result among all strategies, with a mean AUC of 0.6082, outperforming the full-feature baseline and all comparison variants. GSTS also obtained the lowest mean rank and ranked within the top two in every task--label combination, demonstrating that its advantage was not driven by a single paradigm or psychopathology dimension. Importantly, this performance gain was achieved while retaining only a small subset of the original feature space, indicating that granularity-aware selection can remove redundant features without discarding useful multi-scale EEG information. 
The compact GSTS variant provides a complementary view of this effect. Under a strict feature budget of approximately 200 features, GSTS-compact achieved a mean AUC comparable to the full-feature baseline while reducing the feature space by nearly two orders of magnitude. Its adjusted stability was also higher than that of flat stability selection, suggesting that partitioning features by spatial granularity yields more reproducible feature subsets than applying a single global selection constraint. Taken together, Figure~\ref{fig:ec_mw_comparison} and Table~\ref{tab:feature_select_aggregate} support the same conclusion: GSTS provides the best overall feature selection strategy in terms of the primary AUC metric, while also improving feature compactness and selection stability.

\begin{figure}[t]
\centering
\includegraphics[width=0.8\linewidth]{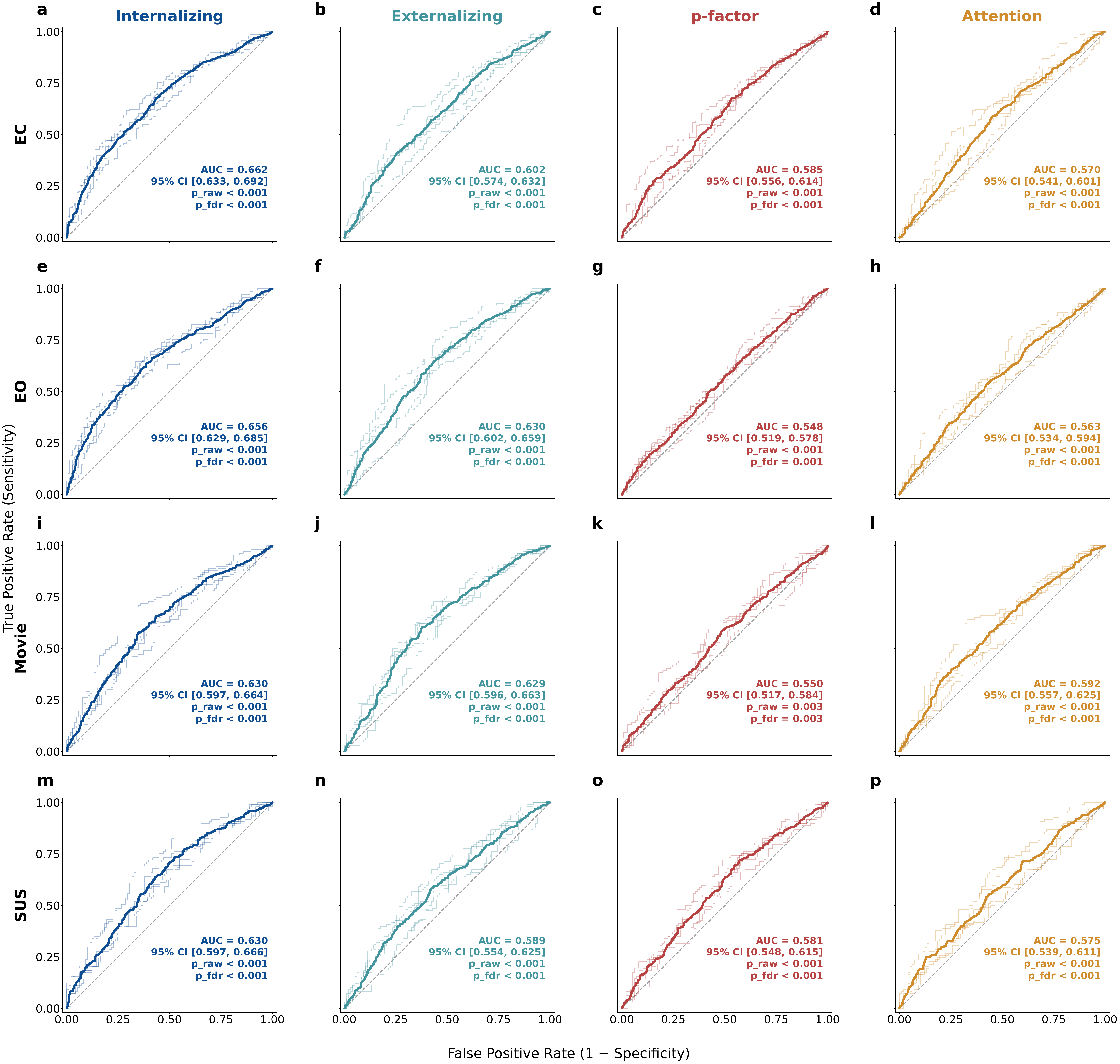}
\caption{\textbf{ROC curves of the proposed GSTS pipeline across all 16 task--label combinations.}
Rows correspond to EEG paradigms (EC, EO, MW, SuS), and columns correspond to psychopathology dimensions (internalizing, externalizing, $p$-factor, attention). Each panel reports AUC with 95\% confidence interval and permutation-based $p$-value against random-chance performance (AUC = 0.5).}
\label{fig:gsts_roc}
\end{figure}

To further verify that the predictive performance of GSTS is meaningfully above random chance rather than a consequence of sampling variability, we next examined the ROC curves and permutation-based significance tests across all task--label combinations. 
Figure~\ref{fig:gsts_roc} shows the ROC curves of GSTS across all 16 task--label combinations. Most panels achieved AUC values above 0.5, and the permutation-based tests confirmed that several task--label combinations were significantly better than random-chance prediction. These results indicate that the signals captured by GSTS are not merely random fluctuations in high-dimensional EEG features. Nevertheless, the AUC values remain moderate, so the results should be interpreted as evidence of weak but detectable EEG correlates of dimensional psychopathology rather than as sufficient performance for individual-level clinical diagnosis.

\begin{figure}[t!]
    \centering
    \includegraphics[width=1\textwidth]{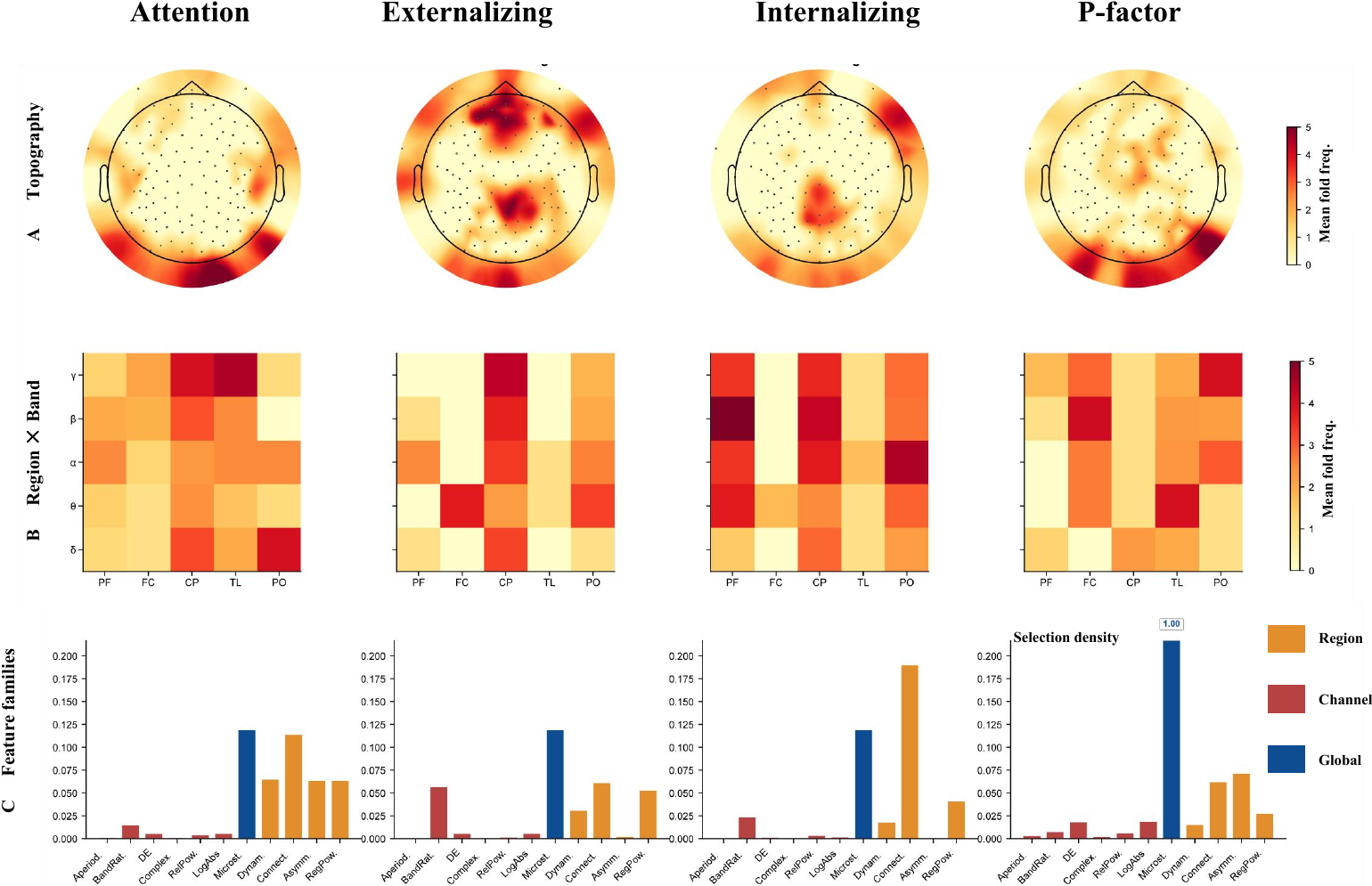}
    \caption{
    \textbf{Interpretable EEG markers identified by the proposed GSTS framework.}
    The figure summarizes stable EEG features selected by the final GSTS pipeline for four psychopathology dimensions: attention, externalizing, internalizing, and $p$-factor.
    (A) Channel-level scalp topographies show the spatial distribution of selection density across electrodes.
    (B) Region-by-frequency heatmaps summarize stable marker distributions across brain regions and canonical frequency bands.
    (C) Feature-family contributions show the selection density of different EEG feature types, with colors indicating spatial granularity: global, region, and channel.
    Selection density was computed as the mean fold-wise selection frequency across outer cross-validation folds.
    }
    \label{fig:interpretable_markers}
\end{figure}

\subsection*{Visualization}

To facilitate interpretation of the selected EEG markers, we visualized the selection density of features aggregated across outer cross-validation folds for each psychopathology dimension, as shown in Figure~\ref{fig:interpretable_markers}. Selection density reflects how frequently a feature was retained across folds and serves as a proxy for the stability of its predictive contribution; it does not imply causal relevance or functional activation. \textbf{Panel A} shows that channel-level selection density exhibited dimension-specific spatial preferences: attention problems focused over posterior and occipitotemporal regions; externalizing showed prefrontal, frontocentral, and midline hotspots; internalizing concentrated over central and right posterior/temporal electrodes; and the $p$-factor displayed a diffuse posterior-dominant pattern. \textbf{Panel B} reveals that selected features concentrated in specific region--frequency-band interactions rather than distributing uniformly, suggesting that predictive information arises from joint spatial--spectral patterns rather than any single band or region in isolation. \textbf{Panel C} indicates that GSTS did not converge on a single preferred granularity but adaptively integrated global, regional, and channel-level features in a target-dependent manner—global microstates most prominent for $p$-factor, region-level features for internalizing, and channel-level features for externalizing—empirically confirming that granularity-aware selection preserves multi-scale information rather than allowing any single spatial level to dominate.

\begin{table}[t]
\centering
\caption{\textbf{Exploratory sanity check of GSTS on the PEARL resting-state EEG dataset (BDI depression prediction).}
Results are reported as AUC / Accuracy / Macro F1 (mean $\pm$ std) under eyes-closed (EC) and eyes-open (EO) conditions. Full: no selection; GSTS: granularity-aware stability selection.}
\label{tab:pearl_phase2_gsts}
\begin{threeparttable}
\renewcommand{\arraystretch}{1.15}
\resizebox{1\linewidth}{!}{%
\begin{tabular}{lcccccc}
\toprule
\textbf{Condition}
& \textbf{Strategy}
& \textbf{AUC}$\uparrow$
& \textbf{Accuracy}$\uparrow$
& \textbf{Macro F1}$\uparrow$
& \textbf{\#Features}

& \textbf{Conclusion} \\
\midrule

\multirow{3}{*}{EC}
& Full (no selection)
& $0.688 \pm 0.214$
& $0.596 \pm 0.141$
& $0.579 \pm 0.153$
& 3{,}426

& Baseline \\

& Dummy
& $0.587 \pm 0.116$
& $0.586 \pm 0.111$
& $0.581 \pm 0.114$
& 3{,}426

& Chance \\

& GSTS (Ours)
& $\mathbf{0.744 \pm 0.217}$
& $\mathbf{0.687 \pm 0.195}$
& $\mathbf{0.664 \pm 0.213}$
& 61

& \textbf{Stress-test pass} \\

\midrule

\multirow{3}{*}{EO}
& Full (no selection)
& $0.486 \pm 0.173$
& $0.505 \pm 0.119$
& $0.488 \pm 0.113$
& 3{,}426

& Baseline \\

& Dummy
& $0.587 \pm 0.116$
& $0.586 \pm 0.111$
& $0.581 \pm 0.114$
& 3{,}426

& Chance baseline \\

& GSTS (Ours) 
& $\mathbf{0.608 \pm 0.288}$
& $\mathbf{0.603 \pm 0.217}$
& $\mathbf{0.597 \pm 0.223}$
& 95

& \textbf{Stress-test pass} \\

\bottomrule
\end{tabular}%
}
\end{threeparttable}
\end{table}

\subsection*{Exploratory Sanity Check on an Independent Adult Cohort (PEARL)}

To probe whether the granularity-aware selection pipeline remains technically operable under protocol shift, we applied the identical feature extraction and GSTS workflow—without adaptation or re-tuning—to the PEARL resting-state EEG dataset~\cite{dzianok2024pearl}. PEARL differs from HBN along every relevant axis: age (50--63 vs.\ 5--21 years), sample size ($n=79$ vs.\ $n=2{,}575$), EEG hardware, and target construct (BDI depression scores vs.\ CBCL-derived psychopathology dimensions). These mismatches are deliberate: publicly available EEG datasets with clinically relevant phenotypes remain scarce, and most candidates suitable for external probing contain approximately 100 participants~\cite{dzianok2024pearl}. We therefore frame this analysis as an exploratory stress test rather than a formal external validation. 
From Table~\ref{tab:pearl_phase2_gsts}, given the small PEARL sample ($n=79$), all comparisons fall within overlapping standard-error intervals and formal statistical inference is unwarranted. Under eyes-closed (EC) resting-state, GSTS yielded an AUC of $0.744 \pm 0.217$ for BDI prediction, compared to $0.688 \pm 0.214$ for the no-selection baseline (Full), while reducing features from 3{,}426 to 61. Under eyes-open (EO), GSTS achieved AUC $0.608 \pm 0.288$ versus Full at $0.486 \pm 0.173$, with feature reduction to 95 features. Cross-fold stability of selected feature subsets was moderate in both conditions (EC: 0.705, EO: 0.779). Notably, in the EO condition, the full feature set (Full) underperformed the dummy-classifier baseline (AUC 0.587), indicating that the raw signal-to-noise ratio was too low for meaningful prediction without feature selection; GSTS surpassed the dummy baseline, suggesting that granularity-aware selection can recover predictive signal even when the unselected feature space is uninformative. 
The results offer a weak but non-negative signal: the GSTS pipeline did not collapse under severe distribution shift, and the HBN gains are unlikely to be trivially explained by overfitting to a single dataset's feature structure.

\subsection*{Clinical Utility Considerations}
Although the AUC values reported in this study (ranging from approximately 0.52 to 0.63) may appear modest compared with those achieved in binary diagnostic classification tasks, their clinical utility should be interpreted in the context of screening and risk stratification rather than definitive individual diagnosis. In real-world pediatric mental health settings, the base rate of clinically significant psychopathology is relatively low, and behavioral checklists alone often yield high false-positive rates due to rater variability and situational factors. Even moderate AUC (~0.60) could be useful for population-level risk stratification, but individual-level diagnosis is not recommended. When used as a triage tool, an EEG-derived risk score can meaningfully reorder referral priority—for example, by identifying the top 30\% of children with the highest predicted $p$-factor or externalizing scores for follow-up clinical assessment, while reducing unnecessary screenings in the low-risk tail. This ``rule-out'' utility, even at moderate discriminative performance, may offer practical value in resource-constrained primary care or school-based mental health screening programs, where objective, low-burden biomarkers are urgently needed to complement subjective ratings. Moreover, the consistency of predictive signals across multiple paradigms (especially EC and MW) suggests that the framework captures stable, replicable neural correlates rather than task-specific artifacts, which is a prerequisite for any clinically deployable tool. We emphasize that these considerations are qualitative and hypothesis-generating; rigorous evaluation of clinical utility would require prospective studies with predefined risk thresholds, decision-curve analysis, and cost-benefit assessment in real-world referral pathways.

\section*{Conclusion}
We present a reproducible computational recipe that systematically integrates multi-paradigm recordings and multi-scale descriptors to predict dimensional psychopathology scores in children and adolescents. Our results indicate that resting-state and naturalistic paradigms carry measurable but weak neural correlates of psychopathology dimensions, with nonlinear tree-based models and granularity-balanced feature selection showing encouraging performance relative to conventional flat-selection or linear approaches in certain conditions. The ablation and visualization analyses suggest that different spatial granularities contribute complementary information, with regional-level features serving as a robust intermediate representation and channel-level features offering additional fine-grained details for specific phenotypes such as attention and externalizing problems. An exploratory sanity check on an independent adult resting-state cohort suggested that the granularity-aware selection pipeline remained technically operable under protocol shifts; this preliminary observation is not intended as evidence of generalizability but rather as a demonstration of pipeline portability, pending rigorous validation on larger, age-matched cohorts. This work offers a reproducible, fold-contained computational pipeline and provides empirical evidence that spatial granularity is a potentially important consideration in EEG-based psychiatric phenotyping, warranting further investigation in future studies.

\section*{Acknowledgments}
\textbf{Funding:} This work is supported by National Key R\&D Program of China (NO.2024YFB3311602), the Ministry of Education of the People's Republic of China (JYB2025XDXM605), Natural Science Foundation of China (62272144, 72188101), the Anhui Provincial Natural Science Foundation (2408085J040), and the Major Project of Anhui Provincial Science and Technology Breakthrough Program (202423k09020001), the Anhui Provincial Graduate Quality Engineering Program (2024cxcysj002), the Fundamental Research Funds for the Central Universities (JZ2024AHST0337), and the New Cornerstone Science Foundation through the XPLORER PRIZE and the CAST Young Talent Cultivation Program for Doctoral Students.

\noindent\textbf{Author contributions:} {Conceptualization:} H.C., J.H., and J.Z.  
{Methodology:} H.C., J.H., J.P., and J.Z.  
{Software:} H.C., S.F., and {S.Q.F.}  
{Investigation:} H.C., J.H., J.P., and {M.S.}  
{Visualization:} H.C., J.H., and J.P.  
{Writing—original draft:} H.C. and J.H.  
{Writing—review and editing:} J.H., M.W., D.G., and J.Z.  
{Supervision:} M.W., D.G., and J.Z.  
{Funding acquisition:} J.H., M.W., D.G., and J.Z.

\noindent\textbf{Competing interests:} The authors declare that they have no competing interests.

\section*{Data Availability}
This study utilized the Healthy Brain Network (HBN) EEG dataset~\cite{shirazi2024hbn}, which is publicly available at \url{https://fcon_1000.projects.nitrc.org/indi/cmi_healthy_brain_network/}. From this dataset, we used four EEG paradigms: two resting-state tasks (eyes-closed and eyes-open) and two naturalistic/task-based paradigms (movie watching and surround suppression). For the exploratory PEARL assessment, we used the PEARL dataset~\cite{dzianok2024pearl}, accessible at \url{https://openneuro.org/datasets/ds004796/versions/1.0.4}. 
The code and feature extraction pipeline used in this study will be released upon publication at \url{https://github.com/chf12581/GAFF}.

\printbibliography

@article{dzianok2024pearl,
  title={PEARL-Neuro Database: EEG, fMRI, health and lifestyle data of middle-aged people at risk of dementia},
  author={Dzianok, Patrycja and Kublik, Ewa},
  journal={Scientific Data},
  volume={11},
  number={1},
  pages={276},
  year={2024},
  publisher={Nature Publishing Group UK London}
}

@article{shirazi2024hbn,
  title={HBN-EEG: The FAIR implementation of the Healthy Brain Network (HBN) electroencephalography dataset},
  author={Shirazi, Seyed Yahya and Franco, Alexandre and Scopel Hoffmann, Maur{\'\i}cio and Esper, Nathalia B and Truong, Dung and Delorme, Arnaud and Milham, Michael P and Makeig, Scott},
  journal={bioRxiv},
  pages={2024--10},
  year={2024},
  publisher={Cold Spring Harbor Laboratory}
}

@article{Welch1967,
    title={The use of fast Fourier transform for the estimation of power spectra: A method based on time averaging over short, modified periodograms},
  author={Welch, Peter},
  journal={IEEE Transactions on audio and electroacoustics},
  volume={15},
  number={2},
  pages={70--73},
  year={1967},
  publisher={IEEE}
}

@article{Buzsaki2006,
   title={Neuronal oscillations in cortical networks},
  author={Buzsaki, Gyorgy and Draguhn, Andreas},
  journal={science},
  volume={304},
  number={5679},
  pages={1926--1929},
  year={2004},
  publisher={American Association for the Advancement of Science}
}

@book{Cohen2014,
    title={Analyzing neural time series data: theory and practice},
  author={Cohen, Mike X},
  year={2014},
  publisher={MIT press}
}

@article{Bastos2016,
   title={A tutorial review of functional connectivity analysis methods and their interpretational pitfalls},
  author={Bastos, Andr{\'e} M and Schoffelen, Jan-Mathijs},
  journal={Frontiers in systems neuroscience},
  volume={9},
  pages={175},
  year={2016},
  publisher={Frontiers Media SA}
}

@article{Monastra2001,
    title={The development of a quantitative electroencephalographic scanning process for attention deficit--hyperactivity disorder: Reliability and validity studies.},
  author={Monastra, Vincent J and Lubar, Joel F and Linden, Michael},
  journal={Neuropsychology},
  volume={15},
  number={1},
  pages={136},
  year={2001},
  publisher={American Psychological Association}
}

@article{Clarke2001,
    title={Excess beta activity in children with attention-deficit/hyperactivity disorder: an atypical electrophysiological group},
  author={Clarke, Adam R and Barry, Robert J and McCarthy, Rory and Selikowitz, Mark},
  journal={Psychiatry research},
  volume={103},
  number={2-3},
  pages={205--218},
  year={2001},
  publisher={Elsevier}
}

@article{Arns2013,
    title={A decade of EEG theta/beta ratio research in ADHD: a meta-analysis},
  author={Arns, Martijn and Conners, C Keith and Kraemer, Helena C},
  journal={Journal of attention disorders},
  volume={17},
  number={5},
  pages={374--383},
  year={2013},
  publisher={Sage Publications Sage CA: Los Angeles, CA}
}

@article{He2010,
    title={The temporal structures and functional significance of scale-free brain activity},
  author={He, Biyu J and Zempel, John M and Snyder, Abraham Z and Raichle, Marcus E},
  journal={Neuron},
  volume={66},
  number={3},
  pages={353--369},
  year={2010},
  publisher={Elsevier}
}

@article{Donoghue2020,
    title={Parameterizing neural power spectra into periodic and aperiodic components},
  author={Donoghue, Thomas and Haller, Matar and Peterson, Erik J and Varma, Paroma and Sebastian, Priyadarshini and Gao, Richard and Noto, Torben and Lara, Antonio H and Wallis, Joni D and Knight, Robert T and others},
  journal={Nature neuroscience},
  volume={23},
  number={12},
  pages={1655--1665},
  year={2020},
  publisher={Nature Publishing Group US New York}
}

@article{Hjorth1970,
   title={EEG analysis based on time domain properties},
  author={Hjorth, Bo},
  journal={Electroencephalography and clinical neurophysiology},
  volume={29},
  number={3},
  pages={306--310},
  year={1970},
  publisher={Elsevier}
}

@article{liang2015eeg,
  title={EEG entropy measures in anesthesia},
  author={Liang, Zhenhu and Wang, Yinghua and Sun, Xue and Li, Duan and Voss, Logan J and Sleigh, Jamie W and Hagihira, Satoshi and Li, Xiaoli},
  journal={Frontiers in computational neuroscience},
  volume={9},
  pages={16},
  year={2015},
  publisher={Frontiers Media SA}
}

@article{Davidson1998,
    title={Anterior electrophysiological asymmetries, emotion, and depression: Conceptual and methodological conundrums},
  author={Davidson, Richard J},
  journal={Psychophysiology},
  volume={35},
  number={5},
  pages={607--614},
  year={1998},
  publisher={Cambridge University Press}
}

@article{Allen2004,
      title={Issues and assumptions on the road from raw signals to metrics of frontal EEG asymmetry in emotion},
  author={Allen, John JB and Coan, James A and Nazarian, Maria},
  journal={Biological psychology},
  volume={67},
  number={1-2},
  pages={183--218},
  year={2004},
  publisher={Elsevier}
}

@article{Lachaux1999,
    title={Measuring phase synchrony in brain signals},
  author={Lachaux, Jean-Philippe and Rodriguez, Eugenio and Martinerie, Jacques and Varela, Francisco J},
  journal={Human brain mapping},
  volume={8},
  number={4},
  pages={194--208},
  year={1999},
  publisher={Wiley Online Library}
}

@article{Nolte2004,
    title={Identifying true brain interaction from EEG data using the imaginary part of coherency},
  author={Nolte, Guido and Bai, Ou and Wheaton, Lewis and Mari, Zoltan and Vorbach, Sherry and Hallett, Mark},
  journal={Clinical neurophysiology},
  volume={115},
  number={10},
  pages={2292--2307},
  year={2004},
  publisher={Elsevier}
}

@article{Vidaurre2018,
   title={Spontaneous cortical activity transiently organises into frequency specific phase-coupling networks},
  author={Vidaurre, Diego and Hunt, Laurence T and Quinn, Andrew J and Hunt, Benjamin AE and Brookes, Matthew J and Nobre, Anna C and Woolrich, Mark W},
  journal={Nature communications},
  volume={9},
  number={1},
  pages={2987},
  year={2018},
  publisher={Nature Publishing Group UK London}
}

@article{jia2024enhancing,
  title={Enhancing brain--computer interface performance by incorporating brain-to-brain coupling},
  author={Jia, Tianyu and Sun, Jingyao and McGeady, Ciar{\'a}n and Ji, Linhong and Li, Chong},
  journal={Cyborg and Bionic Systems},
  volume={5},
  pages={0116},
  year={2024},
  publisher={AAAS}
}

@article{ju2022recognition,
  title={Recognition of drivers’ hard and soft braking intentions based on hybrid brain-computer interfaces},
  author={Ju, Jiawei and Feleke, Aberham Genetu and Luo, Longxi and Fan, Xinan},
  journal={Cyborg and Bionic Systems},
  year={2022},
  publisher={AAAS}
}

@article{bao2023predicting,
  title={Predicting moral elevation conveyed in Danmaku comments using EEGs},
  author={Bao, Chenhao and Hu, Xin and Zhang, Dan and Lv, Zhao and Chen, Jingjing},
  journal={Cyborg and Bionic Systems},
  volume={4},
  pages={0028},
  year={2023},
  publisher={AAAS}
}

@article{qi2024augmented,
  title={Augmented recognition of distracted driving state based on electrophysiological analysis of brain network},
  author={Qi, Geqi and Liu, Rui and Guan, Wei and Huang, Ailing},
  journal={Cyborg and Bionic Systems},
  volume={5},
  pages={0130},
  year={2024},
  publisher={AAAS}
}

@article{ye2024adaptive,
  title={Adaptive spatial--temporal aware graph learning for EEG-based emotion recognition},
  author={Ye, Weishan and Wang, Jiyuan and Chen, Lin and Dai, Lifei and Sun, Zhe and Liang, Zhen},
  journal={Cyborg and Bionic Systems},
  volume={5},
  pages={0088},
  year={2024},
  publisher={AAAS}
}

@article{li2024domain,
  title={A domain generalization and residual network-based emotion recognition from physiological signals},
  author={Li, Junnan and Li, Jiang and Wang, Xiaoping and Zhan, Xin and Zeng, Zhigang},
  journal={Cyborg and Bionic Systems},
  volume={5},
  pages={0074},
  year={2024},
  publisher={AAAS}
}

@article{si2023cross,
  title={Cross-subject emotion recognition brain--computer interface based on fNIRS and DBJNet},
  author={Si, Xiaopeng and He, Huang and Yu, Jiayue and Ming, Dong},
  journal={Cyborg and Bionic Systems},
  volume={4},
  pages={0045},
  year={2023},
  publisher={AAAS}
}

@inproceedings{dong2023approach,
  title={An approach for EEG data augmentation based on deep convolutional generative adversarial network},
  author={Dong, Yuanzhe and Tang, Xi and Tan, Fangning and Li, Qingge and Wang, Yingying and Zhang, Huanqing and Xie, Jun and Liang, Wenyuan and Li, Guanglin and Fang, Peng},
  booktitle={2022 IEEE International Conference on Cyborg and Bionic Systems (CBS)},
  pages={347--351},
  year={2023},
  organization={IEEE}
}

@inproceedings{wang2025enabling,
  title={Enabling Tensor Completion on Single Trial EEG Signals},
  author={Wang, Wenzhi and Zhao, Peijie and Jia, Hao and Zhang, Yanpeng and Jiang, Yinlai and Duan, Feng},
  booktitle={2025 IEEE International Conference on Cyborg and Bionic Systems (CBS)},
  pages={111--116},
  year={2025},
  organization={IEEE}
}

@article{Michel2018,
  title={EEG microstates as a tool for studying the temporal dynamics of whole-brain neuronal networks: a review},
  author={Michel, Christoph M and Koenig, Thomas},
  journal={Neuroimage},
  volume={180},
  pages={577--593},
  year={2018},
  publisher={Elsevier}
}

@article{Koenig2002,
  title={Millisecond by millisecond, year by year: normative EEG microstates and developmental stages},
  author={Koenig, Thomas and Prichep, Leslie and Lehmann, Dietrich and Sosa, Pedro Valdes and Braeker, Elisabeth and Kleinlogel, Horst and Isenhart, Robert and John, E Roy},
  journal={Neuroimage},
  volume={16},
  number={1},
  pages={41--48},
  year={2002},
  publisher={Elsevier}
}

@article{Lehmann1987,
  title={EEG alpha map series: brain micro-states by space-oriented adaptive segmentation},
  author={Lehmann, Dietrich and Ozaki, Hisaki and P{\'a}l, Ivan},
  journal={Electroencephalography and clinical neurophysiology},
  volume={67},
  number={3},
  pages={271--288},
  year={1987},
  publisher={Elsevier}
}

@inproceedings{Duan2013,
  title={Differential entropy feature for EEG-based emotion classification},
  author={Duan, Ruo-Nan and Zhu, Jia-Yi and Lu, Bao-Liang},
  booktitle={2013 6th international IEEE/EMBS conference on neural engineering (NER)},
  pages={81--84},
  year={2013},
  organization={IEEE}
}

@article{2yun2024advances,
  title={Advances, challenges, and prospects of electroencephalography-based biomarkers for psychiatric disorders: a narrative review},
  author={Yun, Seokho},
  journal={Journal of Yeungnam Medical Science},
  volume={41},
  number={4},
  pages={261--268},
  year={2024},
  publisher={Journal of Yeungnam Medical Science}
}

@article{3poldrack2017scanning,
  title={Scanning the horizon: towards transparent and reproducible neuroimaging research},
  author={Poldrack, Russell A and Baker, Chris I and Durnez, Joke and Gorgolewski, Krzysztof J and Matthews, Paul M and Munaf{\`o}, Marcus R and Nichols, Thomas E and Poline, Jean-Baptiste and Vul, Edward and Yarkoni, Tal},
  journal={Nature reviews neuroscience},
  volume={18},
  number={2},
  pages={115--126},
  year={2017},
  publisher={Nature Publishing Group UK London}
}

@article{4jach2020decoding,
  title={Decoding personality trait measures from resting EEG: An exploratory report},
  author={Jach, Hayley K and Feuerriegel, Daniel and Smillie, Luke D},
  journal={Cortex},
  volume={130},
  pages={158--171},
  year={2020},
  publisher={Elsevier}
}

@article{5luo2025multidimensional,
  title={Multidimensional EEG features integration with feature selection strategy for precision diagnosis of depressive disorders},
  author={Luo, Xiaodong and Xu, Yanting and Yan, Zihao and Liu, Wei and Zhou, Bin and Li, Gang and Zhu, Yixia},
  journal={Frontiers in Psychiatry},
  volume={16},
  pages={1624997},
  year={2025},
  publisher={Frontiers Media SA}
}

@article{6london2019artificial,
  title={Artificial intelligence and black-box medical decisions: accuracy versus explainability},
  author={London, Alex John},
  journal={Hastings Center Report},
  volume={49},
  number={1},
  pages={15--21},
  year={2019},
  publisher={Wiley Online Library}
}

@article{14comai2025moving,
  title={Moving toward precision and personalized treatment strategies in psychiatry},
  author={Comai, Stefano and Manchia, Mirko and Bosia, Marta and Miola, Alessandro and Poletti, Sara and Benedetti, Francesco and Nasini, Sofia and Ferri, Raffaele and Rujescu, Dan and Leboyer, Marion and others},
  journal={International Journal of Neuropsychopharmacology},
  volume={28},
  number={5},
  pages={pyaf025},
  year={2025},
  publisher={Oxford University Press US}
}

@article{bosl2025dynamical,
  title={A dynamical systems framework for precision psychiatry},
  author={Bosl, William J and Enlow, Michelle Bosquet and Nelson, Charles A},
  journal={npj Digital Medicine},
  volume={8},
  number={1},
  pages={586},
  year={2025},
  publisher={Nature Publishing Group UK London}
}

@article{new_2,
  title={Adaptive spatiotemporal encoding network for cognitive assessment using resting state EEG},
  author={Sun, Jingnan and Shen, Anruo and Sun, Yike and Chen, Xiaogang and Li, Yunxia and Gao, Xiaorong and Lu, Bai},
  journal={npj Digital Medicine},
  volume={7},
  number={1},
  pages={375},
  year={2024},
  publisher={Nature Publishing Group UK London}
}

@article{new_11,
  title={Personalized functional brain network topography is associated with individual differences in youth cognition},
  author={Keller, Arielle S and Pines, Adam R and Shanmugan, Sheila and Sydnor, Valerie J and Cui, Zaixu and Bertolero, Maxwell A and Barzilay, Ran and Alexander-Bloch, Aaron F and Byington, Nora and Chen, Andrew and others},
  journal={Nature communications},
  volume={14},
  number={1},
  pages={8411},
  year={2023},
  publisher={Nature Publishing Group UK London}
}

@article{guo2024benchmarking,
title={Benchmarking Micro-action Recognition: Dataset, Methods, and Applications},
author={Guo, Dan and Li, Kun and Hu, Bin and Zhang, Yan and Wang, Meng},
journal={IEEE Transactions on Circuits and Systems for Video Technology},
year={2024},
volume={34},
number={7},
pages={6238-6252},
}

@inproceedings{
hu2026see,
title={See the Emotion: A Facial Emoji Proxy Modeling for {EEG} Emotion Recognition},
author={Jingjing Hu and Dan Guo and Haofan Cheng and Zeng ying and Zhan Si and Jinxing Zhou and Meng Wang},
booktitle={Forty-third International Conference on Machine Learning},
year={2026}
}

@inproceedings{hu2025spatialenergyaware,
  title={Spatial-Energy-Aware Dynamic Filtering with Sparse Graph Convolutions for EEG Emotion Recognition},
  author={Hu, J. and Fu, S. and Si, Z. and Chen, J.},
  booktitle={Proceedings of the Annual Meeting of the Cognitive Science Society},
  volume={47},
  year={2025}
}

@article{taoband,
  title={Band-Gated Identity-Disentangled Training for Cross-Subject Auditory Attention Decoding},
  author={Tao, Siying and Hu, Jingjing and Huang, Jinyang and Cui, Fengqi and Liu, Xueliang and Guo, Dan}
}

@article{liu2018survey,
  title={Survey on stability of feature selection},
  author={Liu, Yi},
  journal={Journal of software},
  volume={29},
  number={9},
  pages={2559--2579},
  year={2018},
  publisher={Science Press}
}

\end{document}